\documentclass[journal]{IEEEtran}
\usepackage{amsmath,amsfonts}
\usepackage{algorithmic}
\usepackage{algorithm}
\usepackage{array}
\usepackage{textcomp}
\usepackage{stfloats}
\usepackage{url}
\usepackage{verbatim}
\usepackage{graphicx}
\usepackage{cite}
\usepackage{multirow}
\usepackage{amssymb}
\usepackage{xcolor}
\usepackage{booktabs}
\usepackage{makecell}
\usepackage{xspace}
\usepackage{enumitem}
\ifCLASSOPTIONcompsoc
\usepackage[caption=false, font=normalsize, labelfont=sf, textfont=sf]{subfig}
\else
\usepackage[caption=false, font=footnotesize]{subfig}
\fi

\newcommand{\ie}[1]{{\textit{i.e.}{{#1}}}}
\newcommand{\eg}[1]{{\textit{e.g.}{{#1}}}}

\begin{document}

\title{Hierarchical Graph Pattern Understanding for Zero-Shot VOS}

\author{Gensheng Pei,
	    Fumin~Shen,
            Yazhou~Yao, 
            Tao~Chen,\\
    	Xian-Sheng~Hua (Fellow, IEEE),
	    and~Heng-Tao Shen (Fellow, IEEE)
      
	\thanks{G. Pei, Y. Yao, and T. Chen are with the School of Computer Science and Engineering, Nanjing University of Science and Technology, China.}
	\thanks{F. Shen and H. Shen are with the School of Computer Science and Engineering, University of Electronic Science and Technology of China, China.}
	\thanks{X. Hua is with the DAMO Academy, Alibaba Group, China.}
	
}

\markboth{IEEE TRANSACTIONS ON IMAGE PROCESSING}%
{Shell \MakeLowercase{\textit{et al.}}: A Sample Article Using IEEEtran.cls for IEEE Journals}


\maketitle

\begin{abstract}
    The optical flow guidance strategy is ideal for obtaining motion information of objects in the video. It is widely utilized in video segmentation tasks. However, existing optical flow-based methods have a significant dependency on optical flow, which results in poor performance when the optical flow estimation fails for a particular scene. The temporal consistency provided by the optical flow could be effectively supplemented by modeling in a structural form. This paper proposes a new hierarchical graph neural network (GNN) architecture, dubbed hierarchical graph pattern understanding (HGPU), for zero-shot video object segmentation (ZS-VOS). Inspired by the strong ability of GNNs in capturing structural relations, HGPU innovatively leverages motion cues (\ie, optical flow) to enhance the high-order representations from the neighbors of target frames. Specifically, a hierarchical graph pattern encoder with message aggregation is introduced to acquire different levels of motion and appearance features in a sequential manner. Furthermore, a decoder is designed for hierarchically parsing and understanding the transformed multi-modal contexts to achieve more accurate and robust results. HGPU achieves state-of-the-art performance on four publicly available benchmarks (DAVIS-16, YouTube-Objects, Long-Videos and DAVIS-17). Code and pre-trained model can be found at \url{https://github.com/NUST-Machine-Intelligence-Laboratory/HGPU}.
\end{abstract}

\begin{IEEEkeywords}
Video object segmentation, graph neural network, zero-shot, encoder-decoder, optical flow.
\end{IEEEkeywords}

\section{Introduction}\label{sec:1}
\IEEEPARstart{V}{ideo} object segmentation (VOS) distinguishes the primary foreground objects from the background in every video frame. It has stimulated research on a multitude of related topics. It has stimulated research on a multitude of related topics and has been successfully applied in daily life. The VOS study is mainly performed in the zero-shot and single-shot cases, where their main difference lies in whether the inference stage provides a pixel-level object mask for the first frame. This paper focuses on the challenging zero-shot VOS task, which automatically segments objects of interest without manual interaction.

\begin{figure}[t]
	\centering
	\vspace{-0.4cm}
	\includegraphics[width=1.0\linewidth]{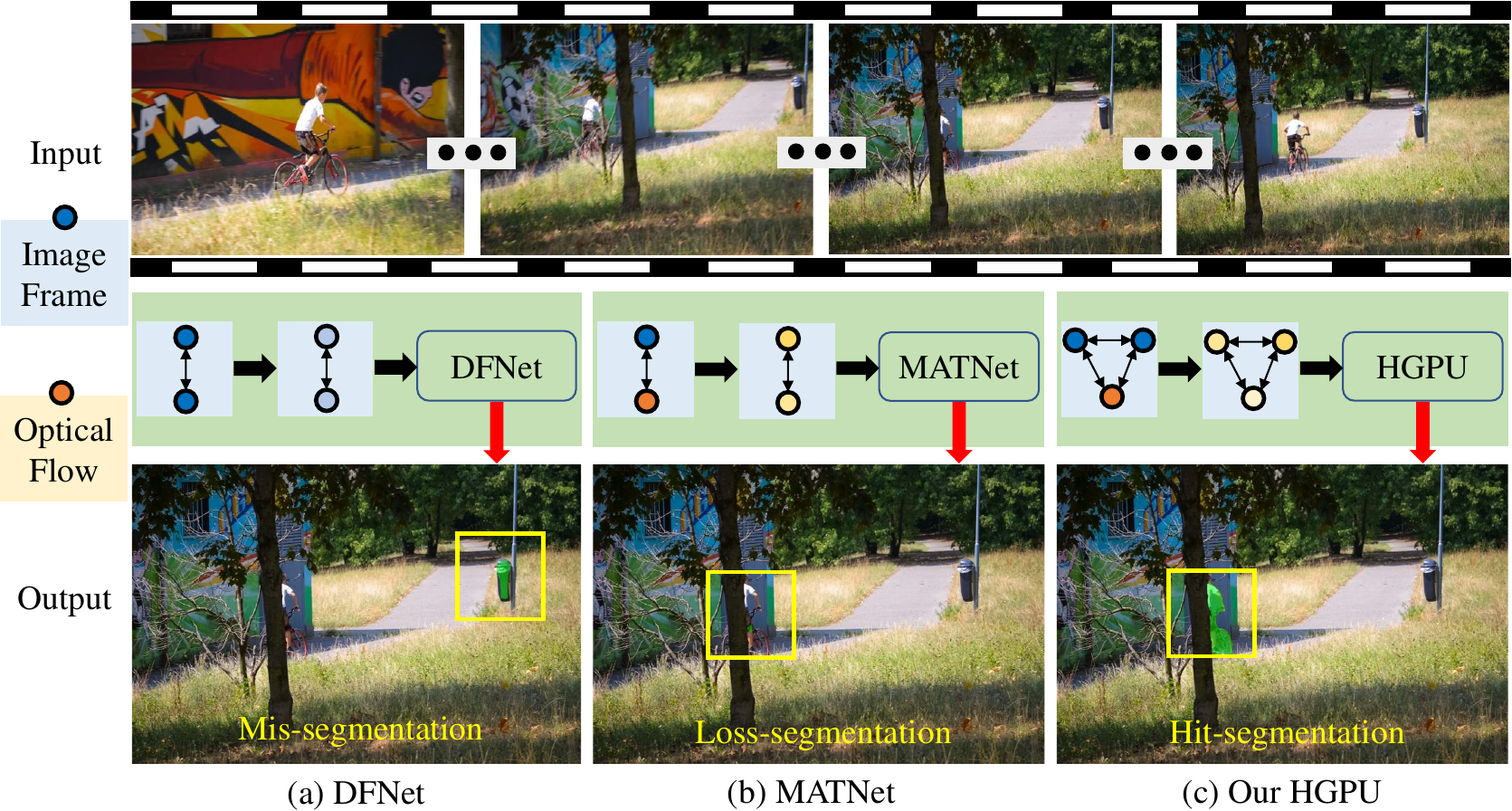}
	\caption{Comparisons of the proposed HGPU and existing methods on \textit{bmx-tress} from DAVIS-16~\cite{perazzi2016benchmark}. (\textbf{a}) DFNet~\cite{zhen2020learning} utilizes appearance features yet without graph reasoning, (\textbf{b}) MATNet~\cite{zhou2020matnet_tip} leverages appearance and optical flow to acquire target regions with motion attention, yet no graph reasoning is performed, (\textbf{c}) Our HGPU constructs the graph on two adjacent frames and their optical flow, which enables their high-order motion-appearance relations to be effectively reasoned for estimating the optimal \textit{primary objects}  in a global-to-local manner. The lost- and mis-segmentation issues on the `bicyclist' are well addressed by HGPU.}
	\label{fig.1}
\end{figure}

Recently, deep learning-based approaches for addressing ZS-VOS have made great strides. Following different feature adoption strategies, the existing models is broadly divided into appearance-based, motion-based and motion-appearance-based. Appearance-based scenarios~\cite{mahadevan2020making,liu2020f2net,zhen2020learning,wang2019zero,yao2021non} draw on well-established image segmentation techniques and video inter-frame relationships. However, the absence of prior information about \textit{primary objects} can lead to mis-segmentation situations in the ZS-VOS manner (Fig. \ref{fig.1} \textbf{(a)}). Further, pure motion information guided approaches~\cite{zhou2021flow,xi2022implicit} transform the ZS-VOS setting into a segmentation task for moving objects. The main drawback of this is the reliance on the motion state of the target and the risk of losing the target when the object moves slowly or is stationary. The natural idea is to synergize appearance and motion features to compensate for the disadvantages of each. Thus, motion-appearance-based methods~\cite{tokmakov2019learning,jain2017fusionseg} can enrich motion information and select appearance features for target regions. However, for video sequences with occlusion and large-scale variation, the motion-appearance features of unstructured modeling inevitably fail to track the \textit{primary objects}, which leads to the lost-segmentation (Fig. \ref{fig.1} \textbf{(b)}).

AGNN \cite{wang2019zero} introduced a technique to mitigate segmentation failures due to object deformation by using graph neural networks (GNNs)\cite{scarselli2008graph} to model high-order correlations among video frames. 
However, mis-segmentation can still occur if backgrounds appear similar to the \textit{primary objects}, as there may be insufficient prior knowledge about the object being segmented.
The high-order relationships extracted by AGNN~\cite{wang2019zero} are limited to appearance, making it challenging to accurately identify the \textit{primary objects} using only appearance features. Nevertheless, their work demonstrated the potential for GNNs to model high-order relationships among contextual frames in zero-shot video object segmentation (ZS-VOS).

To tackle the aforementioned challenges, we propose a hierarchical graph pattern understanding (HGPU) model that captures motion and appearance features from different levels of video nodes. This work follows the current trend in the computer vision field of leveraging optical flow estimation to capture efficient motion information~\cite{zhou2020matnet_tip,AMC-Net,RTNet,TransportNet,pei2023HCPN}. The intuition behind using motion features is that they can guide the representation of the primary object's appearance to overcome variations and noisy backgrounds, and that motion patterns can effectively highlight the main objects and exclude the background. Specifically, HGPU builds a hierarchical representation by comprehensively leveraging motion and appearance features in a unified message propagation framework, which leads to high-quality segmentation results (Fig. \ref{fig.1} \textbf{(c)}).

While relying solely on video appearance features may seem elegant, it is undeniable that these methods pose challenges in identifying reliable primary object discrimination cues in the ZS-VOS task. Optical flow is a well-established method for estimating motion information, and with the rapid development of deep learning, its performance has been significantly improved. Optical flow estimation based on depth methods has more reliable motion capture and extraction capabilities compared to traditional motion feature acquisition methods, such as SIFT and Kalman filter. However, existing optical flow-based methods have a significant dependency on optical flow, which results in poor performance when the estimation of optical flow fails for a particular scene. To address this issue, we propose HGPU, which constructs higher-order forms of motion and appearance features to obtain more robust feature representations. Our approach relies on the robust structured feature representation for excellent performance in the face of multiple video challenge tasks. Moreover, we investigate the potential of this approach in addressing specific challenges that arise in zero-shot video object segmentation.

\begin{figure}[t]
	\begin{center}
		\includegraphics[width=0.97\linewidth]{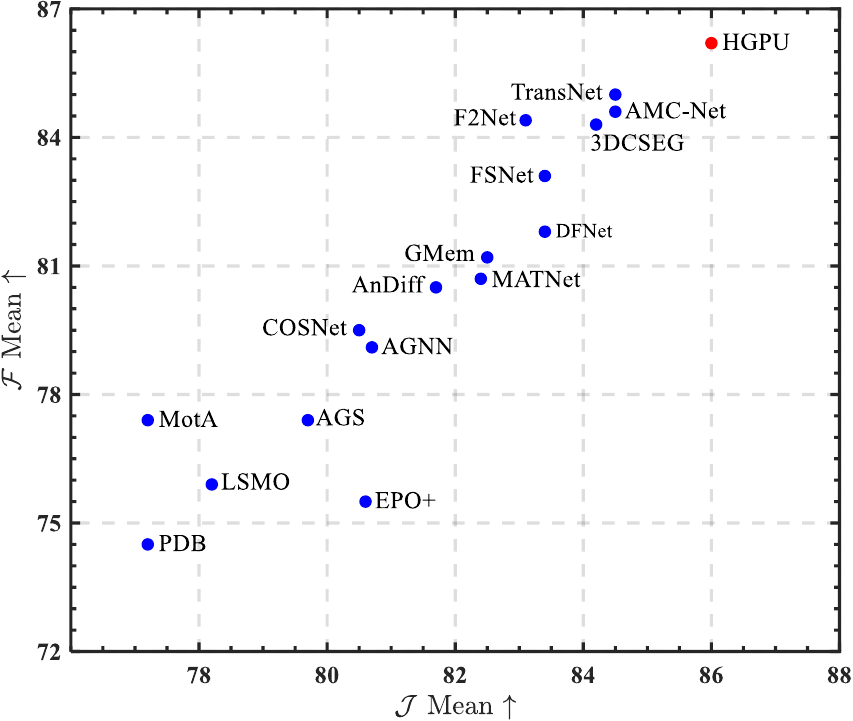}
	\end{center}
	\caption{Boundary accuracy ($\mathcal{F}$) and region similarity ($\mathcal{J}$) of each model on DAVIS-16~\cite{perazzi2016benchmark}. Previous methods and the proposed HGPU are marked as {\large\textcolor[rgb]{0,0,1}{$\bullet$}} and {\large\textcolor[rgb]{1,0,0}{$\bullet$}}, respectively. Our approach realizes the new state-of-the-art.}
	\label{fig.2}
\end{figure}

In conclusion, the main contributions of our work can be summarized as follows:

\begin{itemize}
  \item[1)] Our proposed approach, the Hierarchical Graph Pattern Understanding framework (HGPU), constructs a reciprocal pattern of appearance and motion features using GNNs. This approach enables the two frames of appearance features to collaborate in exploring the object region in cases where optical flow is ineffective. Additionally, when the appearance features are distorted, the motion guidance from the optical flow assists the model in identifying motion-specific appearance features.
  
  \item[2)] We introduce the intuitions of using motion features lies in that they can guide the appearance representation of the primary object to overcome noisy backgrounds and variation. Besides, the target’s motion patterns are effective cues for highlighting the \textit{primary objects} and excluding the background. 

  \item[3)] We conduct exhaustive experiments and ablation studies to investigate the intrinsic mechanism of HGPU. Our model achieves state-of-the-art results on all four benchmark datasets and the best-reported results in object- and instance-level ZS-VOS tasks. Fig. \ref{fig.2} compares the results of our proposed method with those of existing methods on the DAVIS-16 \cite{perazzi2016benchmark} dataset in terms of $\mathcal{F}$ and $\mathcal{J}$ Mean.
\end{itemize}

\section{Related Work}\label{Sec.2}
Previous video object segmentation can be broadly classified as zero-shot (\ie, unsupervised) or one-shot (\ie, semi-supervised). These differ in whether they provide ground-truth pixel-level annotations for the first frame during testing.

\subsection{One-Shot Video Object Segmentation}
One-shot VOS is a spatio-temporal matching problem that involves finding the most similar pixels of an object template constructed in the reference frames. The task uses objects annotated in the first frame to locate them in subsequent query frames. Approaches to one-shot VOS can be broadly classified into four categories: propagation-based, detection-based, hybrid-based, and matching-based models.

Propagation-based approaches \cite{xu2022reliable,ventura2019rvos,wang2018semi} use temporal information to learn an object mask propagator to refine the mask propagated from the previous frame. Some self-supervised VOS \cite{yang2021self,lu2020learning} or few-shot VOS \cite{chen2021delving} methods also use this approach for inference. Detection-based approaches \cite{caelles2017one,lin2021query,ge2021video} create an object detector from the object appearance of the first frame, then crop out the object for segmentation. By integrating the two strategies mentioned above, the hybrid-based approaches \cite{wang2019ranet,wang2021swiftnet} have both the advantages of propagation and detection schemes. Matching-based methods \cite{yang2020collaborative,lu2020video} typically train a typical Siamese matching network to find the pixel that best matches between the past frame(s) and the current frame (or query frame), and then accomplishes the corresponding assignment of labels. Some matching-based works \cite{lu2020video,tokmakov2017learning} use a memory mechanism to previously segmented frame information to learn long-term spatial-temporal information and facilitate learning the evolution of objects over time. The difficulty of one-shot VOS is memorizing the first frame's target area to track these pixels. This setup yields good results, but providing first-frame annotations for video is not applicable in many scenarios.

\begin{figure*}[t]
	\centering
	\includegraphics[width=0.95\linewidth]{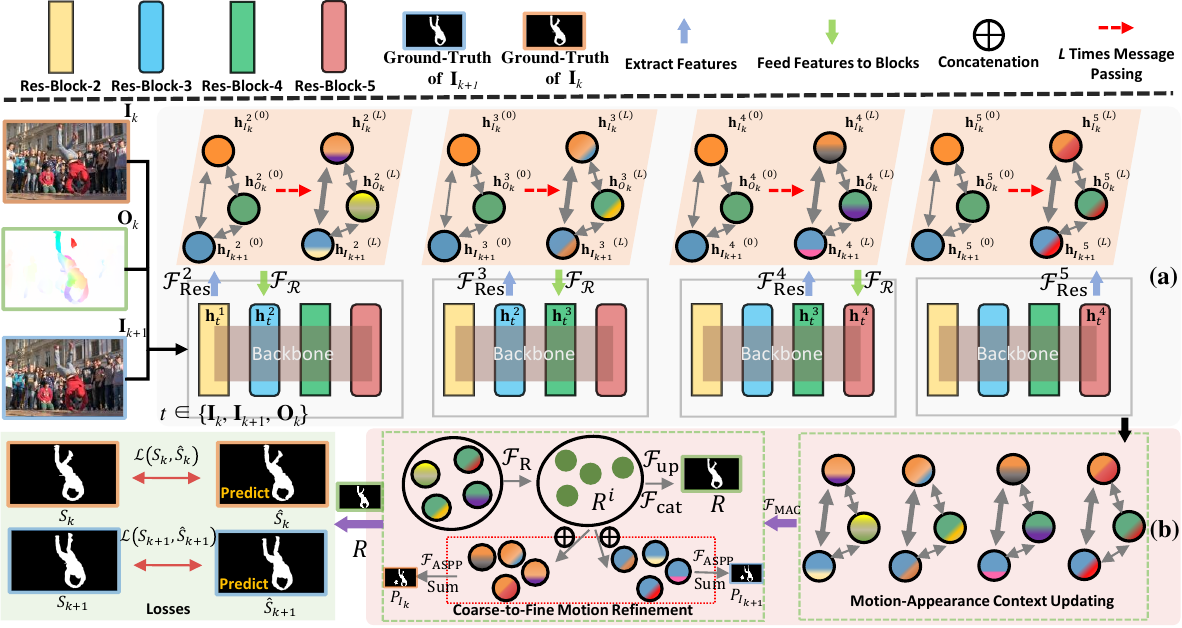}
	\caption{Illustration of HGPU architecture. \textbf{(a) HGPE} (see \S\ref{Sec.3.1} for details) utilizes frames ($I_{k}$, $I_{k+1}$) and their corresponding optical flow ($O_k$) to create high-order feature representations with multiple modalities. This is achieved through the \textit{Message Aggregation}, \textit{Node Update}, and \textit{Motion-Appearance Readout} modules, which hierarchically and structurally represent frames and optical flows. The motion-appearance features are then parsed through \textbf{(b) MAUD} (see \S\ref{Sec.3.2} for details), including the \textit{Motion-Appearance Context Updating} and \textit{Coarse-to-Fine Motion Refinement} modules. Efficient updating and refining of the hierarchical motion-appearance features produce coarse-to-fine segmentation results. The hierarchical GNN framework facilitates the interactive fusion of motion and appearance features for video feature modeling.}\label{fig.3}
\end{figure*}

\subsection{Zero-Shot Video Object Segmentation} 
For a considerable period, the computer vision community has been extensively investigating the issue of automatic binary video segmentation \cite{hampapur1994digital, wang1998unsupervised}. Early conventional models \cite{perazzi2015fully, giordano2015superpixel} relied on specific heuristics such as foregrounds, motion boundaries, and salient objects, which required manual feature extraction. With the development of computational resources and artificial neural networks, deep-based models \cite{ yang2019anchor, song2018pyramid, wang2019learning} have rapidly advanced the ZS-VOS field. LSMO \cite{tokmakov2019learning} is a prime example of the attempt to apply deep learning techniques in this field, which uses a multilayer perceptron to detect moving objects. However, a fully connected network leads to high computational costs and the inability to model spatial information. Hence, many fully convolutional network-based approaches have emerged, such as two-stream networks \cite{li2019motion, siam2019video,pei2022HFAN}, lightweight networks \cite{zhao2021real, fan2022bidirectionally}, and encoder-decoder frameworks \cite{RTNet, AMC-Net, zhou2020matnet_tip, pei2023HCPN}. Siamese networks \cite{lu2019see, lu2020zero, shen2021distilled} and graph neural networks \cite{wang2019zero, lu2020video} extract higher-order information from multiple frames in a video to better understand the video and obtain more accurate foreground estimates. Additionally, various visual attention methods \cite{wang2019learning, wang2020paying, zhou2020matnet_tip, pei2023HCPN} have yielded good performance by taking inspiration from the human visual system.

Some of the problems associated with ZS-VOS are somewhat alleviated by the above-mentioned approaches. However, previous methods that relied solely on inter-frame unstructured features for mining may misidentify the background as the primary object or discard the primary objects in the presence of noise. To address these issues, we propose using GNN to model the inter-frame motion and appearance information. Our proposed approach represents a new paradigm for structured representation of motion and appearance features that is designed explicitly for ZS-VOS.

\subsection{Graph Neural Networks}
Graph neural networks~\cite{scarselli2008graph,yin2021graph} learn feature representations that have strong discriminative capabilities by considering the structural information latent in the data. Hierarchical graph neural networks (HGNNs)~\cite{chen2021hierarchical,xing2021learning,wu2020learning} are effective at capturing the hierarchical structures present in many real-world graphs, such as social networks, protein structures, and neural networks. By incorporating multiple layers of nodes and edges, HGNNs can learn representations that capture both local and global dependencies in the graph. For the first time, in the ZS-VOS task, AGNN~\cite{wang2019zero} are used to construct the structural connectivity of \textit{primary objects} based on GNNs. This approach extracts video inter-frame features and creates a fully connected graph by considering each video frame as a node. AGNN~\cite{wang2019zero} and GMem~\cite{lu2020video} have shown remarkable results in capturing the continuity of moving objects in videos.

However, a structured representation of video frames alone does not address the temporal consistency of the \textit{primary objects}. To address this issue, we propose a structural modeling approach based on video frames that carry motion information to cope with the challenges of ZS-VOS. Additionally, we explore the potential of this approach in addressing specific challenges in ZS-VOS.

\section{The Proposed Approach}\label{Sec.3}
In this section, we introduce the proposed method to address zero-shot video object segmentation based on graph neural networks.
As shown in Fig.~\ref{fig.3}, the proposed HGPU consists of two key modules: \textbf{1)} a hierarchical graph pattern encoder (HGPE), which provides a hierarchical pattern representation of the structural features among video frames (\S\ref{Sec.3.1}), and \textbf{2)} a motion-appearance understanding decoder (MAUD) for parsing and refining the motion-appearance features (\S\ref{Sec.3.2}).

\subsection{Overview and Notations} 
Consider an input video $\mathcal{I}=\{I_{k}\in\mathbb{R}^{w\times{h}\times{3}}\}_{k=1}^N$ with $N$ frames, where $w\times{h}$ is the spatial size of a frame. Its optical flow set is $\mathcal{O}=\{O_{k}\in\mathbb{R}^{w\times{h}\times{3}}\}_{k=1}^{N-1}$ using balanced performance RAFT~\cite{teed2020raft} (RGB images after visualization). The ultimate goal of ZS-VOS is to autonomously derive the binary masks $\mathcal{S}=\{S_k\in\{0,1\}^{w\times{h}}\}_{k=1}^N$ for the video. 

To achieve this demand, HGPU models the video $\mathcal{I}$ as a fully connected graph $\mathcal{G=(V,E)}$, where node $v_t\in\mathcal{V}$ indicates the $t^{\texttt{th}}$ frame and edge $e_{t,u}=(v_t,v_u)\in\mathcal{E}$ represents the association from node $v_t$ to $v_u$. Similar to~\cite{zhou2020matnet_tip}, our HGPU adopts five convolutional blocks of ResNet-101~\cite{he2016deep} as the backbone network to extract optical flow features $O_k$ from neighboring frames $I_k$ and $I_{k+1}$, after the $i^{\texttt{th}}|_{i=2}^5$ residual blocks. In each residual block, the multi-level features are mined for inter-frame high-order correlations in a global-to-local manner. The decoder for understanding motion-appearance features yields the target predictions $\hat{\mathcal{S}}$. In the following, we will specify each of the two main components of the proposed approach.

\subsection{Hierarchical Graph Pattern Encoder}
\label{Sec.3.1}

To improve the region confidence of the target object in a video, we take the optical flow between adjacent frames as motion cues. In the $i^{\texttt{th}}$ residual block, the features are extracted by taking the outputs of the $(i-1)^{\texttt{th}}$ block as inputs, $i\in{\{2,3,4,5\}}$. Specifically, for node $v_t|t\in{\{I_k,I_{k+1},O_k\}}$, the inputs of the $i^{\texttt{th}}$ block and outputs (after graph reasoning) are denoted as $h_{t}^{i-1}$ and $h_{t}^{i}$, respectively. As such, for node $v_t$ of the $i^{\texttt{th}}$ block, we obtain the initial state (outputs of the $(i-1)^{\texttt{th}}$ block) as follows:
\begin{equation}\label{Eq.1}
	\begin{split}
		&h_{t}^i=\mathcal{F}_{\texttt{Res}}^i\left( h_{t}^{i-1}\right)\in\mathbb{R}^{W\times{H}\times{C}}, \\
		&{h_{t}^i}^{(0)}=\mathcal{F}_{\texttt{SA}}\left(h_t^i\right)\in\mathbb{R}^{W\times{H}\times{C}},
	\end{split}
\end{equation}
where $\mathcal{F}_{\texttt{SA}}$ is the soft attention (SA) that emphasizes content-informative regions. ${h_{t}^i}^{(0)}$ is an initial state with a spatial size $W\times{H}$ and channel $C$, corresponding to node $v_{t}$. When $i=1$, $h_t^i$ indicates the original features of $I_k$, $I_{k+1}$, or $O_k$.

\subsubsection{Message Aggregation} After obtaining the initial state for each node ${h_{t}^i}^{(0)}$, inspired by~\cite{nguyen2018improved}, we concentrate on each node's feature representations in parallel to perform message aggregation (see Fig. \ref{fig.3} \textbf{(a)}) between different nodes. To calculate the edge $e_{t,u}|(u\in\{I_k,O_k,I_{k+1}\},t\ne{u})$ under the $l^{\texttt{th}}|_{l=0}^L$ iteration ($L$ denotes the number of message passing and, in this paper, is set to 1), we have
\begin{equation}\label{Eq.2}
	\begin{split}
		{S_{t,u}^i}^{(l)} &= {h_t^i}^{(l)} X \left({h_u^i}^{(l)}\right)^{\texttt{T}} = {h_t^i}^{(l)} \mathcal{P}(X) \left({h_u^i}^{(l)}\right)^{\texttt{T}},\\
		{e_{t,u}^i}^{(l)} &= {S_{t,u}^i}^{(l)}\in\mathbb{R}^{(WH)\times{(HW)}}, \\
		{e_{u,t}^i}^{(l)} &= \left({S_{t,u}^i}^{(l)}\right)^{\texttt{T}}\in\mathbb{R}^{(WH)\times{(HW)}},
	\end{split}
\end{equation}
where $X\in\mathbb{R}^{C\times{C}}$ is the learnable weight matrix, and $\mathcal{P}$ maps $X$ linearly to two low-dimensional matrices to ease the computational complexity. $S$ is the affinity matrix between node features, so the pairwise correlation ${e_{t,u}^i}^{(l)}$ for node $v_t$ to node $v_u$ can be represented by ${S_{t,u}^i}^{(l)}$ (and vice versa for transposing ${S_{t,u}^i}^{(l)}$). As in~\cite{gilmer2017neural}, we design a learnable differential message passing function for the aggregation process. A message transmission from node $v_u$ to node $v_t$ is formulated as follows:
\begin{equation}\label{Eq.3}
	\begin{split}
		{m_{u,t}^{i+1}}^{(l)}&=\mathcal{M}\left({h_t^i}^{(l)},{h_u^i}^{(l)},{e_{t,u}^i}^{(l)}\right),\quad u\in\mathcal{N}_t,\\
		&= {\texttt{softmax}}\left({e_{t,u}^i}^{(l)}\right){h_u^i}^{(l)}\in\mathbb{R}^{W\times{H}\times{C}},
	\end{split}
\end{equation}
where $\mathcal{M}(\cdot,\cdot,\cdot)$ is the message function used to collect messages from a node and its neighbor, and $\mathcal{N}_t$ indicates the neighbors of node $v_t$ in graph $\mathcal{G}$. ${m_{u,t}^{i+1}}^{(l)}$ means that messages are passed from neighboring node $v_u$ to $v_t$, acting as a message aggregator.

\subsubsection{Node Update} Following~\cite{wang2019zero,lu2020video}, we utilize ConvGRU~\cite{ballas2015delving} as a node state's update function to transmit and reinforce the previous state. Note that the states in this paper are executed in the residual blocks (see Fig. \ref{fig.3} (a)). Moreover, for node $v_t$, its updated state is a set of the neighboring nodes' states to preserve the discrepancy of neighboring nodes with respect to node $v_t$. The discriminative motion-appearance fusion features are then passed into the node update,
\begin{equation}\label{Eq.4}
	\begin{split}
		{U_{u,t}^{i+1}}^{(l)}=\mathcal{U}_{\texttt{ConvGRU}}\left({h_t^i}^{(l)},{m_{u,t}^{i+1}}^{(l)}\right),\quad u\in\mathcal{N}_t,
	\end{split}
\end{equation}
where ${U_{u,t}^{i+1}}^{(l)}\in\mathbb{R}^{W\times{H}\times{C}}$ denotes the feature representation of node $v_t$ updated with neighbor information. The update function $\mathcal{U}_{\texttt{ConvGRU}}$ is embedded in each residual block, enabling the update and delivery of messages with motion-appearance-specific features.

\subsubsection{Motion-Appearance Readout} 
\begin{figure*}[t]
	\begin{center}
		\includegraphics[width=\linewidth]{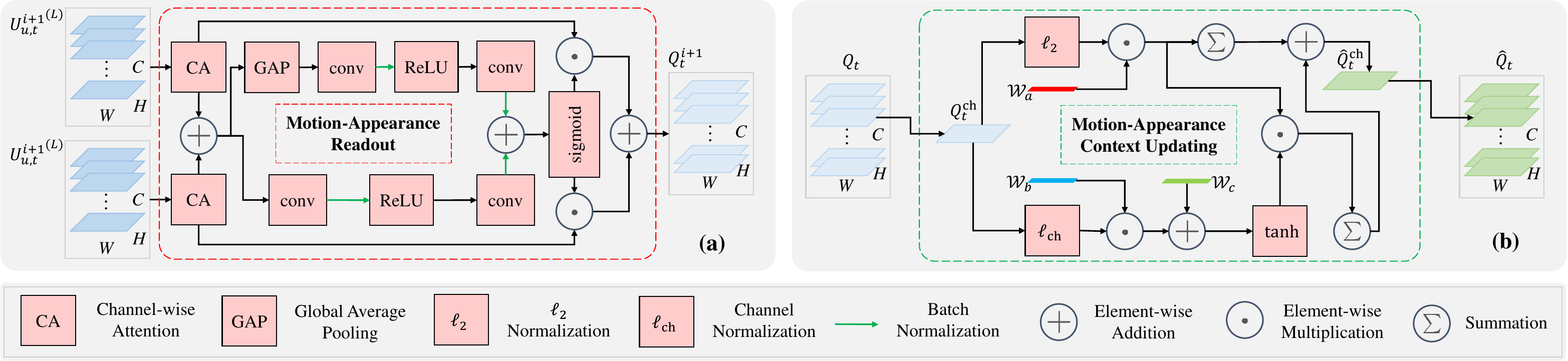}
	\end{center}
	\caption{A detailed illustration of two components: (\textbf{a}) Motion-Appearance Readout (MAR) and (\textbf{b}) Motion-Appearance Context Updating (MACU). MAR performs feature fusion for each node, and MACU creates competing and cooperative associations in appearance and motion cues.}
	\label{fig.4}
\end{figure*}

Unlike the existing GNN-based works~\cite{wang2019zero,lu2020video}, our readout stage does not output the final segmentation predictions, but instead gives each node's feature fusion, which is then passed to the next block for sequential processing. To fuse the features coming from the neighboring nodes and pass them into the next residual block, this paper uses motion-appearance readout (MAR) (Fig. \ref{fig.4} (\textbf{a})) for the integration summary of messages. First, channel-level relations are obtained for each neighboring message using channel attention~\cite{chen2017sca}. Then the global and local motion-appearance feature representations are obtained by $\mathcal{F}_{\texttt{global}}$ and $\mathcal{F}_{\texttt{local}}$, respectively,
\begin{equation}\label{Eq.5}
	\begin{split}
		&\mathcal{F}_{\texttt{global}}\left({U_{u,t}^{i+1}}^{(L)}\right)=\texttt{SeqOp}\left({\texttt{gap}}\left({U_{u,t}^{i+1}}^{(L)}\right)\right), \\
		&\mathcal{F}_{\texttt{local}}\left({U_{u,t}^{i+1}}^{(L)}\right)=\texttt{SeqOp}\left({U_{u,t}^{i+1}}^{(L)}\right), \\
	\end{split}
\end{equation}
where $\texttt{SeqOp}(\cdot)$ is achieved by $\texttt{conv}_{1\times{1}}\rightarrow\texttt{BN}\rightarrow\texttt{ReLU}\rightarrow\texttt{conv}_{1\times{1}}\rightarrow\texttt{BN}$. Here, $\texttt{gap}$ is the global average pooling, $\texttt{BN}$ indicates the batch normalization, and $\texttt{ReLU}$ represents the rectified linear unit. For node $v_t$, our motion-appearance readout produces an embedding summarization $Q_t^{i+1}$ with greater expressive power than simply accumulating the final node states,
\begin{equation}\label{Eq.6}
	\begin{split}
	    \mathcal{F}_{\mathcal{R}}(\cdot)&={\texttt{sigmoid}}\left(\mathcal{F}_{\texttt{global}}(\cdot)+\mathcal{F}_{\texttt{local}}(\cdot)\right) \\
		h_t^{i+1}&=Q_t^{i+1}=\mathcal{F}_{\mathcal{R}}\left({U_{u,t}^{i+1}}^{(L)}\right), \quad u\in\mathcal{N}_t,
	\end{split}
\end{equation}
where $h_t^{i+1}\in\mathbb{R}^{W\times{H}\times{C}}$ represents the state of node $v_t$ in the $(i+1)^{\texttt{th}}$ residual block. $\mathcal{F}_{\mathcal{R}}$ serves as the interaction center for messages received from neighboring nodes. It also enhances and merges the features they convey to make them available for the next residual block.

\subsection{Motion-Appearance Understanding Decoder}\label{Sec.3.2}
Rather than directly applying an iterative state update to the last layer of the backbone network, a sophisticated sequence of operations (Message Aggregation $\rightarrow$ Node Update $\rightarrow$ Motion-Appearance Readout) is performed between the residual blocks, without requiring iterations. Hierarchical reasoning over the graph pattern representation of the video enables the acquisition of higher-order motion-appearance features from a global-to-local perspective.

\subsubsection{Motion-Appearance Context Updating} Appearance features compensate for the absence of motion descriptions on the object semantics, while motion information enables selecting high-quality regions for appearance features. Inspired by~\cite{yang2020gated}, we apply a contextual association-based approach to create associations of competition and cooperation in appearance and motion cues. This work performs a channel-level enhancement of the motion-appearance context updating (MACU) (Fig. \ref{fig.4} (\textbf{b})). To summarize the $Q_t$ (the superscript $i$ is removed for simplicity) of the node $v_t$, the motion-appearance context is
\begin{equation}\label{Eq.7}
	\begin{split}
        \hat{Q}_t&=\mathcal{F}_{\texttt{MACU}}\left(Q_t,|\mathcal{W}_a,\mathcal{W}_b,\mathcal{W}_c\right)\in\mathbb{R}^{W\times{H}\times{C}}, \\
		&=\sum_{{\texttt{ch}}=1}^{C}\mathcal{W}_a\ell_{2}\left(Q_t^{\texttt{ch}}\right)\times\\
		&\left(1+{\texttt{tanh}}\left(\mathcal{W}_b\ell_{\texttt{channel}}\left(Q_t^{\texttt{ch}}\right)+\mathcal{W}_c\right)\right),
	\end{split}
\end{equation}
where $\mathcal{W}_a$, $\mathcal{W}_b$, $\mathcal{W}_c\in\mathbb{R}^C$ are the learnable weight parameters. $\ell_2$ and $\ell_{\texttt{channel}}$ indicate $\ell_2$-normalization and channel-normalization, respectively. $\mathcal{F}_{\texttt{MACU}}$ acts as a relay station to augment the feature discrimination of $Q_t$ from HGPE, outputting a tensor $\hat{Q}_t$ of the same size.

\subsubsection{Coarse-to-Fine Motion Refinement} Since the \textit{primary objects} are often the ones that stand out, their motion patterns can provide crucial clues for segmentation. We leverage the motion information carried by the optical flow to determine the target objects and refine the coarse segmentation. For this purpose, we establish a motion reference with node $v_t|t=O_k$:
\begin{equation}\label{Eq.8}
	\begin{split}
		&\mathcal{F}_{\texttt{R}}(\cdot)={\texttt{sigmoid}}\left({\texttt{conv}}_{1\times{1}}({\texttt{conv}}_{3\times{3}}(\cdot))\right), \\
		&R^i=\mathcal{F}_{\texttt{R}}\left(Q_{t}^i\right), R_{\texttt{up}}^i={\mathcal{F}_{\texttt{up}}}(R^i),\\
		&R_{\texttt{concat}}={\mathcal{F}_{\texttt{concat}}}\left(\{R_{\texttt{up}}^i\}_{i=2}^5\right), \\
		&R={\texttt{sigmoid}}\left({\mathcal{F}_{\texttt{pred}}}\left(R_{\texttt{concat}}\right)\right)\in[0,1]^{w\times{h}},
	\end{split}
\end{equation}
where $\mathcal{F}_{\texttt{R}}(\cdot)$ uses motion features to construct the contour reference of the \textit{primary targets}. $\mathcal{F}_{\texttt{up}}$ and $\mathcal{F}_{\texttt{concat}}$ indicate bilinear upsampling and concatenation, respectively. $\mathcal{F}_{\texttt{pred}}$ maps $R_{\texttt{concat}}\in[0,1]^{w\times{h}\times{4}}$ to $\mathbb{R}\in[0,1]^{w\times{h}}$ with ${\texttt{conv}}_{3\times{3}}$. The final motion reference $R$ is acquired by $\texttt{sigmoid}$. 
To enlarge the receptive field and obtain sufficient spatial information, following~\cite{zhou2020matnet_tip}, the feature maps concatenated between $\hat{Q}_{t}^i$ and $R^i$ are fed into $\mathcal{F}_{\texttt{ASPP}}$~\cite{chen2017deeplab} to yield a multi-scale representation, \ie,
\begin{equation}\label{Eq.9}
	\begin{split}
		P_{t}^i &= {\mathcal{F}_{\texttt{concat}}}\left(\hat{Q}_{t}^i,R^i\right), t\in{\{I_k,I_{k+1}\}}, \\
		P_{t} &= \sum_{i=2}^{5}\mathcal{F}_{\texttt{ASPP}}\left(P_{t}^i\right),
	\end{split}
\end{equation}
where $P_{t=I_k}$ and $P_{t=I_{k+1}}\in[0,1]^{w\times{h}}$ denote the coarse segmentation results of two consecutive frames. Based on the motion reference $R$ (see Fig. \ref{fig.3} \textbf{(b)}, coarse-to-fine motion refinement (CFMR) block), the fine-grained segmentation of $k^{\texttt{th}}$ frame is 
\begin{equation}\label{Eq.10}
	\begin{split}
		\hat{\mathcal{S}}_k = &P_k\odot{R}, \\
		\hat{\mathcal{S}} = \{\hat{\mathcal{S}}_k\in&{[0,1]^{w\times{h}}}\}_{k=1}^{N} ,
	\end{split}
\end{equation}
where $\odot$ indicates the Hadamard product. The CFMR module helps to segment the results from coarse to fine and outputs the final segmentation $\hat{\mathcal{S}}$ of the input video. Following common practice~\cite{maninis2018deep,zhou2020matnet_tip}, our model is trained to minimize the loss function $\mathcal{L}$ (see Fig. \ref{fig.3} \textbf{(b)}, losses block),
\begin{equation}\label{Eq.11}
	\begin{split}
		\mathcal{L}\left(\mathcal{S},\hat{\mathcal{S}}\right)=-\sum\nolimits_{p}\mathcal{W}_p\mathcal{L}_{\texttt{BCE}}\left(\mathcal{S},\hat{\mathcal{S}}\right), 
	\end{split}
\end{equation}
where $\mathcal{L}_{\texttt{BCE}}$ represents the binary cross-entropy loss, and $\mathcal{W}_p$ denotes the weights assigned to the hard pixel $p$ within a frame. This balanced loss has been found to perform well in boundary detection~\cite{xie2015holistically,maninis2017convolutional,arani2021rgpnet} due to the presence of an overwhelming background portion in the frame.

\begin{table*}[t]
    \centering
    \caption{Quantitative evaluation on the validation set of DAVIS-16~\cite{perazzi2016benchmark} (see \S{\ref{subsec:5.2}} for details). FPS denotes frame per second for inference speed on a V100 GPU. FLOPs indicates floating point operations. The top three performances are marked in \textcolor{red}{red}, \textcolor{blue}{blue} and \textcolor{green}{green} for each metric, respectively. These notes are the same for Tables \ref{Tab.2}, \ref{Tab.4}.}\label{Tab.1}
    \resizebox{18.15cm}{!}{
    \begin{tabular}{r|c|c|ccc|ccc|c|c|c|c}
	\hline
	\multirow{2}*{Method~~~~~~} &\multirow{2}*{Publication} &\multirow{2}*{Backbone} &\multicolumn{3}{c|}{$\mathcal{J}$} &\multicolumn{3}{c|}{$\mathcal{F}$} &$\mathcal{T}$ &{$\mathcal{J}$\&$\mathcal{F}$} &\multirow{2}*{FPS $\uparrow$} &\multirow{2}*{FLOPs $\downarrow$}\\
	~ & & &Mean $\uparrow$ &Recall $\uparrow$ &Decay $\downarrow$ &Mean $\uparrow$ &Recall $\uparrow$ &Decay $\downarrow$ &Mean $\downarrow$ &Mean $\uparrow$ &~ &~\\
	\hline
	LVO~\cite{tokmakov2017learning} &ICCV 2017 &ResNet-101 &75.9 &89.1 &\textcolor{red}{0.0} &72.1 &83.4 &1.3 &26.5 &74.0 &- &-\\
	PDB~\cite{song2018pyramid} &ECCV 2018 &ResNet-50 &77.2 &90.1 &\textcolor{green}{0.9} &74.5 &84.4 &\textcolor{red}{-0.2} &29.1 &75.9 &\textcolor{red}{20.0} &-\\
	LSMO~\cite{tokmakov2019learning} &IJCV 2019 &ResNet-101 &78.2 &89.1 &4.1 &75.9 &84.7 &3.5 &21.2 &77.1 &- &-\\
	MotA~\cite{siam2019video} &ICRA 2019 &ResNet-101 &77.2 &87.8 &5.0 &77.4 &84.4 &3.3 &27.9 &77.3 &- &-\\
	AGS~\cite{wang2019learning} &CVPR 2019 &ResNet-101 &79.7 &91.1 &1.9 &77.4 &85.8 &1.6 &26.7 &78.6 &1.7 &-\\
	AGNN~\cite{wang2019zero} &ICCV 2019 &ResNet-101 &80.7 &94.0 &\textcolor{red}{0.0} &79.1 &90.5 &\textcolor{blue}{0.0} &33.7 &79.9 &1.9 &\textcolor{green}{76.9G}\\
	COSNet~\cite{lu2019see} &CVPR 2019 &ResNet-101 &80.5 &93.1 &4.4 &79.5 &89.5 &5.0 &18.4 &80.0 &2.2 &\textcolor{blue}{75.6G}\\
	AnDiff~\cite{yang2019anchor} &ICCV 2019 &ResNet-101 &81.7 &90.9 &2.2 &80.5 &85.1 &\textcolor{green}{0.6} &21.4 &81.1 &2.8 &-\\
	MATNet~\cite{zhou2020matnet_tip} &TIP 2020 &ResNet-101 &82.4 &94.5 &3.8 &80.7 &90.2 &4.5 &21.6 &81.5 &1.3 & 98.6G\\
  	EPO+~\cite{akhter2020epo} &WACV 2020 &ResNet-101 &80.6 &95.2 &2.2 &75.5 &87.9 &2.4 &19.3 &78.1 &- &-\\
	GMem~\cite{lu2020video} &ECCV 2020 &ResNet-50 &82.5 &94.3 &4.2 &81.2 &90.3 &5.6 &19.8 &81.9 &\textcolor{green}{5.0} &87.5G\\
	DFNet~\cite{zhen2020learning} &ECCV 2020 &ResNet-101 &83.4 &94.4 &4.2 &81.8 &89.0 &3.7 &\textcolor{red}{15.2} &82.6 &3.6 &80.7G\\
	3DCSeg~\cite{mahadevan2020making} &BMVC 2020 &ResNet-152 &84.2 &95.8 &7.4 &84.3 &92.4 &5.5 &\textcolor{green}{16.8} &84.2 &4.5 &-\\
        F2Net~\cite{liu2020f2net} &AAAI 2021 &ResNet-101 &83.1 &95.7 &\textcolor{red}{0.0} &84.4 &92.3 &0.8 &20.9 &83.7 &\textcolor{blue}{10.0} &\textcolor{red}{64.8G}\\
	RTNet~\cite{RTNet} &CVPR 2021 &ResNet-101 &85.6 &\textcolor{green}{96.1} &- &84.7 &\textcolor{blue}{93.8} &- &- &85.2 &0.7 &180.7G\\
         APS~\cite{zhao2021multi} &MM 2021 &ResNet-101 &83.3 &- &- &82.1 &- &- &- &82.7 &- & - \\
	TransNet~\cite{TransportNet} &ICCV 2021 &ResNet-101 &84.5 &- &- &\textcolor{green}{85.0} &- &- &- &84.8 &3.6 &101.3G\\
	AMC-Net~\cite{AMC-Net} &ICCV 2021 &ResNet-101 &84.5 &\textcolor{red}{96.4} &2.8 &84.6 &\textcolor{blue}{93.8} &2.5 &- &84.6 &- & 82.0G \\
        DBSNet~\cite{fan2022bidirectionally} &MM 2022 &ShuffleNet &\textcolor{blue}{85.9} &- &- &84.7 &- &- &- &\textcolor{green}{85.3} &- & - \\
        HCPN~\cite{pei2023HCPN} &TIP 2023 &ResNet-101 &\textcolor{green}{85.8} &\textcolor{blue}{96.2} &3.4 &\textcolor{blue}{85.4} &\textcolor{green}{93.2} &3.0 &18.6 &\textcolor{blue}{85.6} &0.9 & 93.3G\\
	\hline
	\textbf{Ours} &- &ResNet-101 &\textcolor{red}{86.0} &\textcolor{green}{96.1} &5.7 &\textcolor{red}{86.2} &\textcolor{red}{94.1} &4.3 &\textcolor{blue}{16.0}  &\textcolor{red}{86.1} &1.0 & 93.5G\\
	\hline
    \end{tabular}
    }
\end{table*}

\subsection{Implementation Details}
HGPU is an end-to-end learnable model. The input of the model consists of video frames and their corresponding optical flows, all resized to 512$\times$512 resolution.

\subsubsection{Training Phase} 
We pre-train our method on a large dataset before fine-tuning it on the target video dataset.

\textbf{Pre-training:} 
We follow the protocol of the top-performing methods~\cite{zhou2020matnet_tip} and pre-trained our network on YouTube-VOS, comprising a training set of 3,945 video sequences and a validation set of 474 video sequences, to extract more discriminative features directly from the real video data.

\textbf{Main-training:}
For object-level tasks, we fine-tune the proposed model on the training set of DAVIS-16 benchmark~\cite{perazzi2016benchmark}. To optimize the model for instance-level segmentation, they used 60 sequences from the DAVIS-17 dataset~\cite{pont2017} to fine-tune the model pre-trained on YouTube-VOS.

\textbf{Instance-level:}
Inspired by~\cite{garg2021mask}, we develop a foreground mask selection method using space-time memory (STM) as a framework, as proposed by~\cite{oh2019video}. We first generate instance-level masks for the first frame of each validation video using Mask R-CNN~\cite{he2017mask}. Our method, HGPU, is used to produce object-level masks for each frame to select the foreground from redundant proposals. We then utilize a merging approach~\cite{luiten2020unovost} to combine the instance masks obtained by STM with the object masks acquired by our method for achieving the final instance-level results.

\subsubsection{Inference Phase}
After training, we resize the video frames and optical flow resolution to 512$\times$512 and input them into the trained model for segmenting unseen videos. We only apply CRF~\cite{krahenbuhl2011efficient} post-processing for the DAVIS-16 results, following common practice~\cite{wang2019zero,zhou2020matnet_tip,lu2020video,pei2023HCPN,fan2022bidirectionally}. For the remaining benchmarks, we use the network output directly. The inference of HGPU includes optical flow estimation and mask prediction, with image writing. It takes approximately 0.30s and 0.25s per frame, respectively. Post-processing takes about 0.50s per frame.

\begin{table*}[t]
	\centering
	\caption{Quantitative evaluation on the test set of YouTube-Objects \cite{prest2012learning} over $\mathcal{J}$ Mean (see \S{\ref{subsec:5.2}} for details). ``Avg.'' indicates the mean result of the 10 categories.}\label{Tab.2}
		\begin{tabular}{c|ccccccccccccc}
			\hline
			\multirow{2}*{Method}  &MotA &LVO &FSEG &PDB &AGS &COSNet &AGNN &MATNet &GMem &RTNet &AMC-Net &TMO &Ours \\
			~ &~\cite{siam2019video} &~\cite{tokmakov2017learning} &~\cite{jain2017fusionseg} &~\cite{song2018pyramid} &~\cite{wang2019learning} &~\cite{lu2019see} &~\cite{wang2019zero} &~\cite{zhou2020matnet_tip} &~\cite{lu2020video} &~\cite{RTNet} &~\cite{AMC-Net} &~\cite{TMO} &\\
			\hline
			Airplane &77.2 &\textcolor{green}{86.2} &81.7 &78.0 &\textcolor{blue}{87.7} &81.1 &81.1 &72.9 &86.1 &84.1 &78.9 &85.7 &\textcolor{red}{89.7} \\
			Bird &42.2 &\textcolor{blue}{81.0} &63.8 &80.0 &76.7 &75.7 &75.9 &77.5 &75.7 &80.2 &\textcolor{green}{80.9} &80.0 &\textcolor{red}{84.6} \\
			Boat &49.3 &68.5 &\textcolor{red}{72.3} &58.9 &\textcolor{blue}{72.2} &\textcolor{green}{71.3} &70.7 &66.9 &68.6 &70.1 &67.4 &70.1 &69.8 \\
			Car &68.6 &69.3 &74.9 &76.5 &78.6 &77.6 &78.1 &79.0 &\textcolor{red}{82.4} &\textcolor{green}{79.5} &\textcolor{blue}{82.0} &78.0 &75.1 \\
			Cat &46.3&58.8 &68.4 &63.0 &69.2 &66.5 &67.9 &\textcolor{red}{73.7} &65.9 &\textcolor{green}{71.8} &69.0 &\textcolor{blue}{73.6} &67.6 \\
			Cow &64.2 &68.5 &68.0 &64.1 &64.6 &69.8 &69.7 &67.4 &\textcolor{blue}{70.5} &70.1 &69.6 &\textcolor{green}{70.3} &\textcolor{red}{71.7} \\
			Dog &66.1 &61.7 &69.4 &70.1 &73.3 &\textcolor{green}{76.8} &\textcolor{red}{77.4} &75.9 &\textcolor{blue}{77.1} &71.3 &75.8 &\textcolor{green}{76.8} &75.8 \\
			Horse &64.8 &53.9 &60.4 &\textcolor{green}{67.6} &64.4 &67.4 &67.3 &63.2 &\textcolor{red}{72.2} &65.1 &63.0 &66.2 &\textcolor{blue}{69.5} \\
			Motorbike &44.6 &60.8 &62.7 &58.4 &62.1 &\textcolor{blue}{67.7} &\textcolor{red}{68.3} &62.6 &63.8 &64.6 &63.4 &58.6 &\textcolor{green}{65.1} \\
			Train &42.3 &\textcolor{blue}{66.3} &\textcolor{green}{62.2} &35.3 &48.2 &46.8 &47.8 &51.0 &47.8 &53.3 &57.8 &47.0 &\textcolor{red}{70.5} \\
			\hline
                \textbf{FPS} &- &- &- &\textcolor{red}{20.0} &1.7 &\textcolor{green}{2.2} &1.9 &1.3 &\textcolor{blue}{5.0} &0.7 &- &-  &1.0 \\
			\textbf{Avg.} &58.1 &67.5 &68.4 &65.5 &69.7 &70.5 &70.8 &69.0 &\textcolor{blue}{71.4} &71.0 &71.1 &\textcolor{green}{71.5}  &\textcolor{red}{73.9} \\
			\hline
	\end{tabular}
\end{table*}

\begin{table*}[t]	
	\centering
	\caption{Quantitative evaluation on the validation set of Long-Videos~\cite{liang2020video} (see \S{\ref{subsec:5.2}} for details). The top four models are OS-VOS solutions, and the bottom four approaches are under the ZS-VOS setting. The best results are marked in \textbf{bold}.}\label{Tab.3}
		\begin{tabular}{r|c|ccc|ccc|c|c}
			\hline
			\multirow{2}*{Method~~~~~} &\multirow{2}*{Supervision} &\multicolumn{3}{c|}{$\mathcal{J}$} &\multicolumn{3}{c|}{$\mathcal{F}$} &$\mathcal{J}$\&$\mathcal{F}$ &\multirow{2}*{FPS}\\
			~ &~ &Mean $\uparrow$ &Recall $\uparrow$ &Decay $\downarrow$ &Mean $\uparrow$ &Recall$\uparrow$ &Decay $\downarrow$ &Mean $\uparrow$ & \\
			\hline
			RVOS~\cite{ventura2019rvos} &\multirow{4}*{OS-VOS} &10.2 &6.7 &13.0 &14.3 &11.7 &\textbf{10.1} &12.2 &\textbf{22.7} \\
			A-GAME~\cite{johnander2019generative} &~  &50.0 &58.3 &39.6 &50.7 &58.3 &45.2 &50.3 &15.0\\
			STM~\cite{oh2019video}  &~ &79.1 &88.3 &11.6 &79.5 &90.0 &15.4 &79.3 &3.4\\
			AFB-URR~\cite{liang2020video} &~  &\textbf{82.7} &\textbf{91.7} &\textbf{11.5} &\textbf{83.8} &\textbf{91.7} &13.9 &\textbf{83.3} &4.0\\
			\hline
			3DCSeg~\cite{mahadevan2020making} &\multirow{4}*{ZS-VOS} &34.2 &38.6 &11.6 &33.1 &28.1 &15.6 &33.7 &\textbf{4.5} \\
			MATNet~\cite{zhou2020matnet_tip}  &~ &66.4 &73.7 &10.9 &69.3 &77.2 &10.6 &67.9 &1.3 \\
			AGNN~\cite{wang2019zero} &~  &68.3 &77.2 &13.0 &68.6 &77.2 &16.6 &68.5 &1.9 \\
			\textbf{Ours} &~  &\textbf{74.0} &\textbf{82.5} &\textbf{9.2} &\textbf{79.8} &\textbf{91.2} &\textbf{-2.0} &\textbf{76.9} &1.0 \\
			\hline
	\end{tabular}
\end{table*}

\begin{figure*}[t]
	\begin{center}
		\includegraphics[width=1\linewidth]{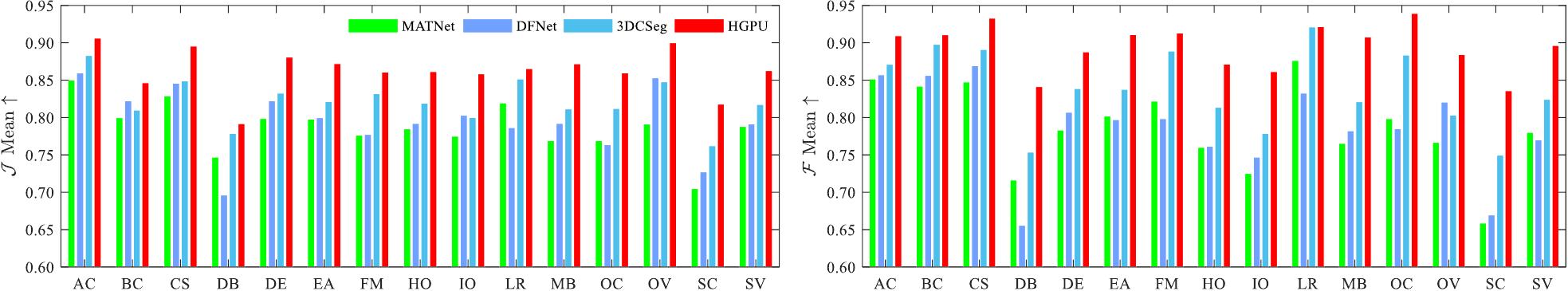}
	\end{center}
	\caption{Attribute-based analysis of the DAVIS-16~\cite{perazzi2016benchmark} val-set. $\mathcal{J}$ and $\mathcal{F}$ Mean are scored  overall sequences with specified attributes. Zoom in for best view.}\label{fig.5}
\end{figure*}

\begin{figure*}
	\centering
	\subfloat[breakdance]{\includegraphics[width=0.5\linewidth]{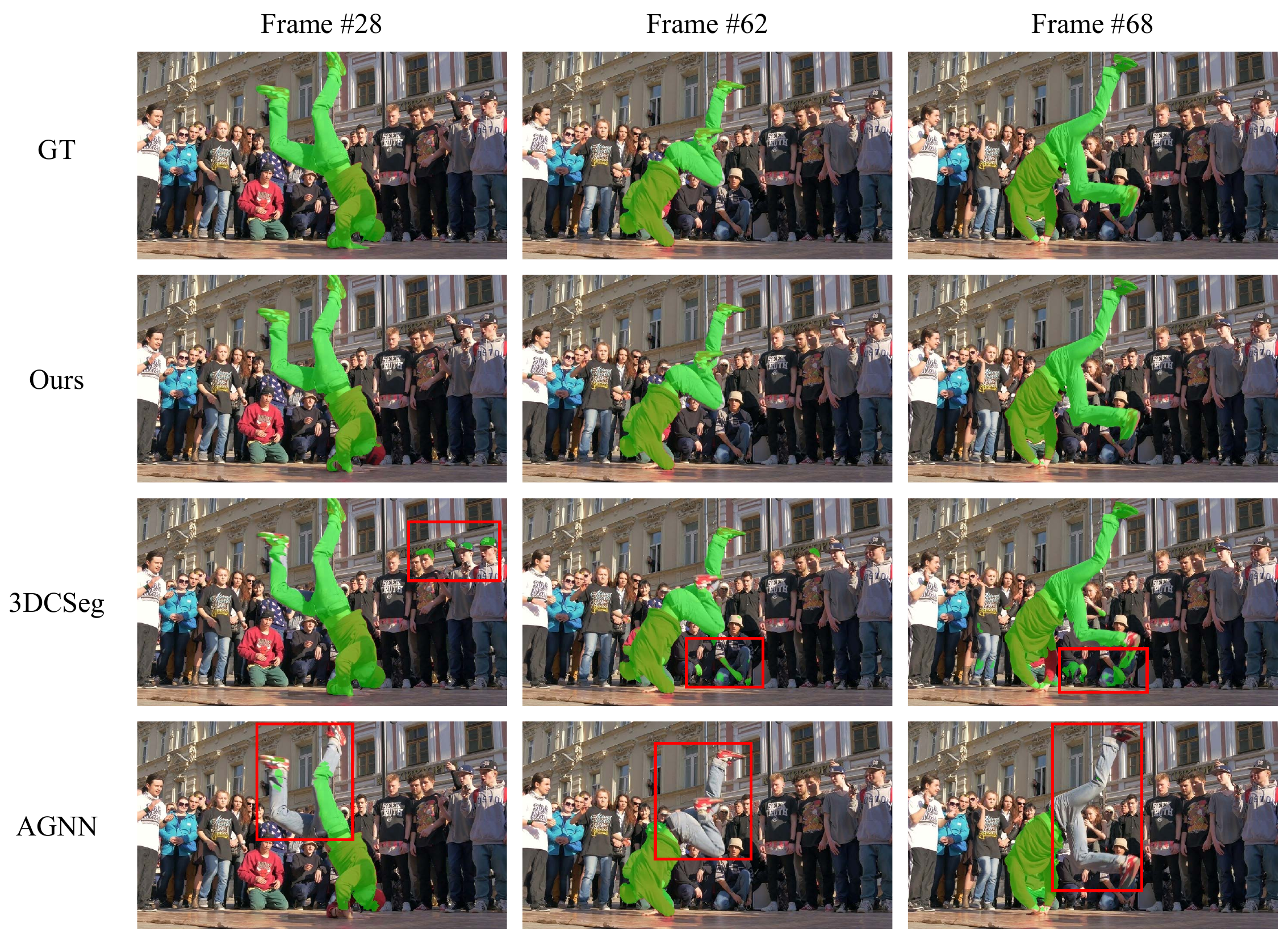}}
	\subfloat[scooter-black]{\includegraphics[width=0.5\linewidth]{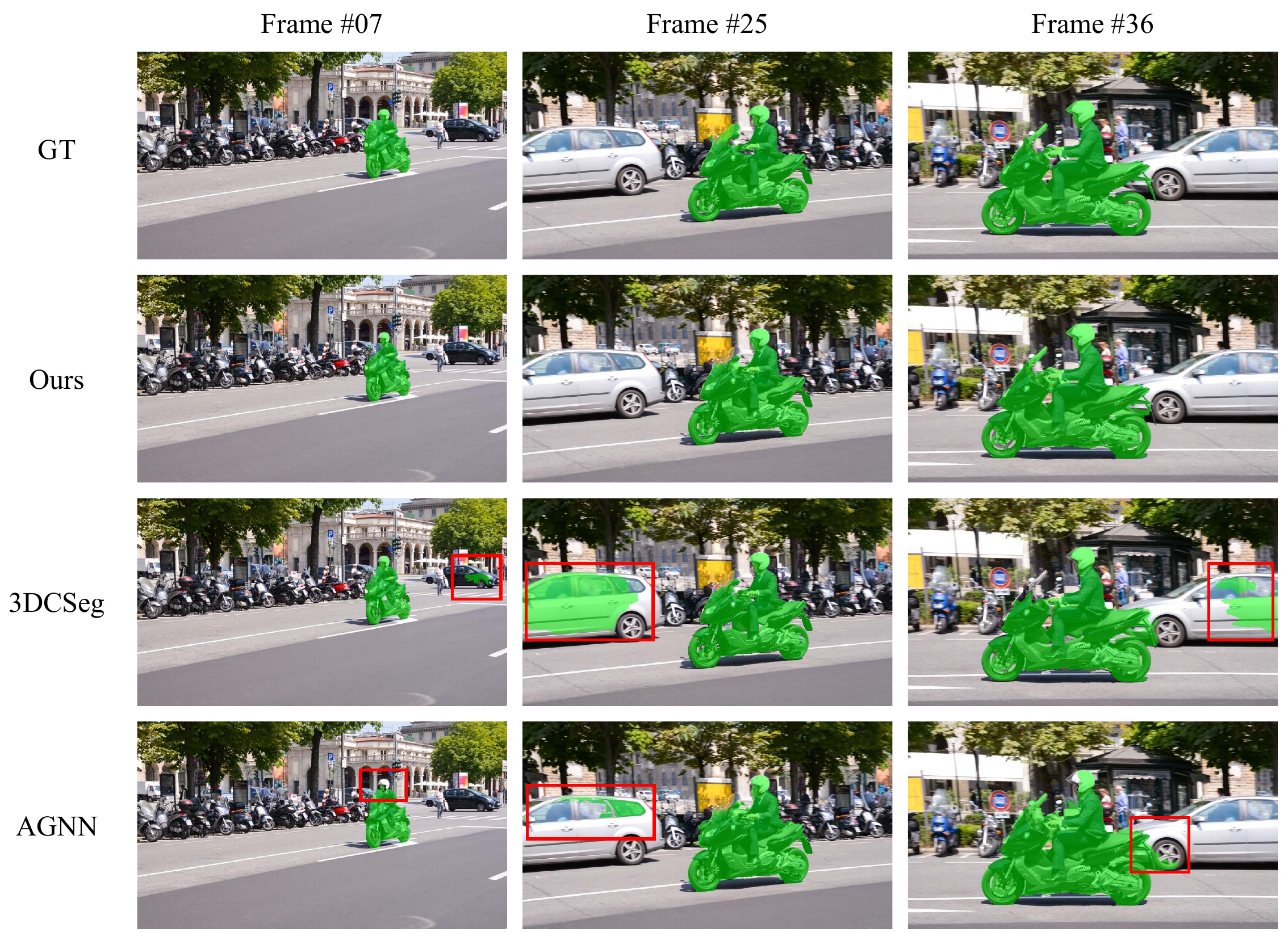}}\\
	\subfloat[train0024]{\includegraphics[width=0.5\linewidth]{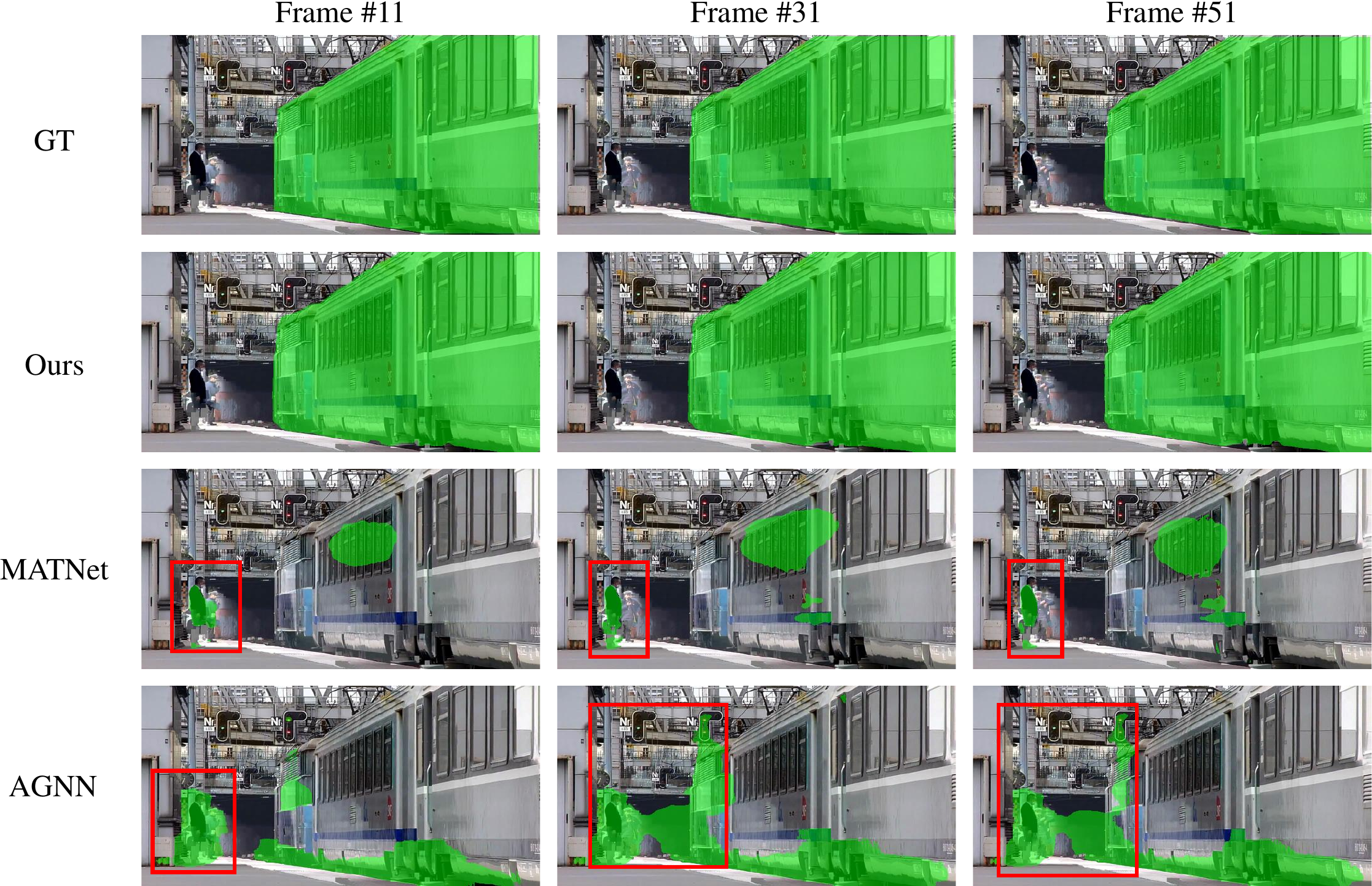}}
	\subfloat[rat]{\includegraphics[width=0.5\linewidth]{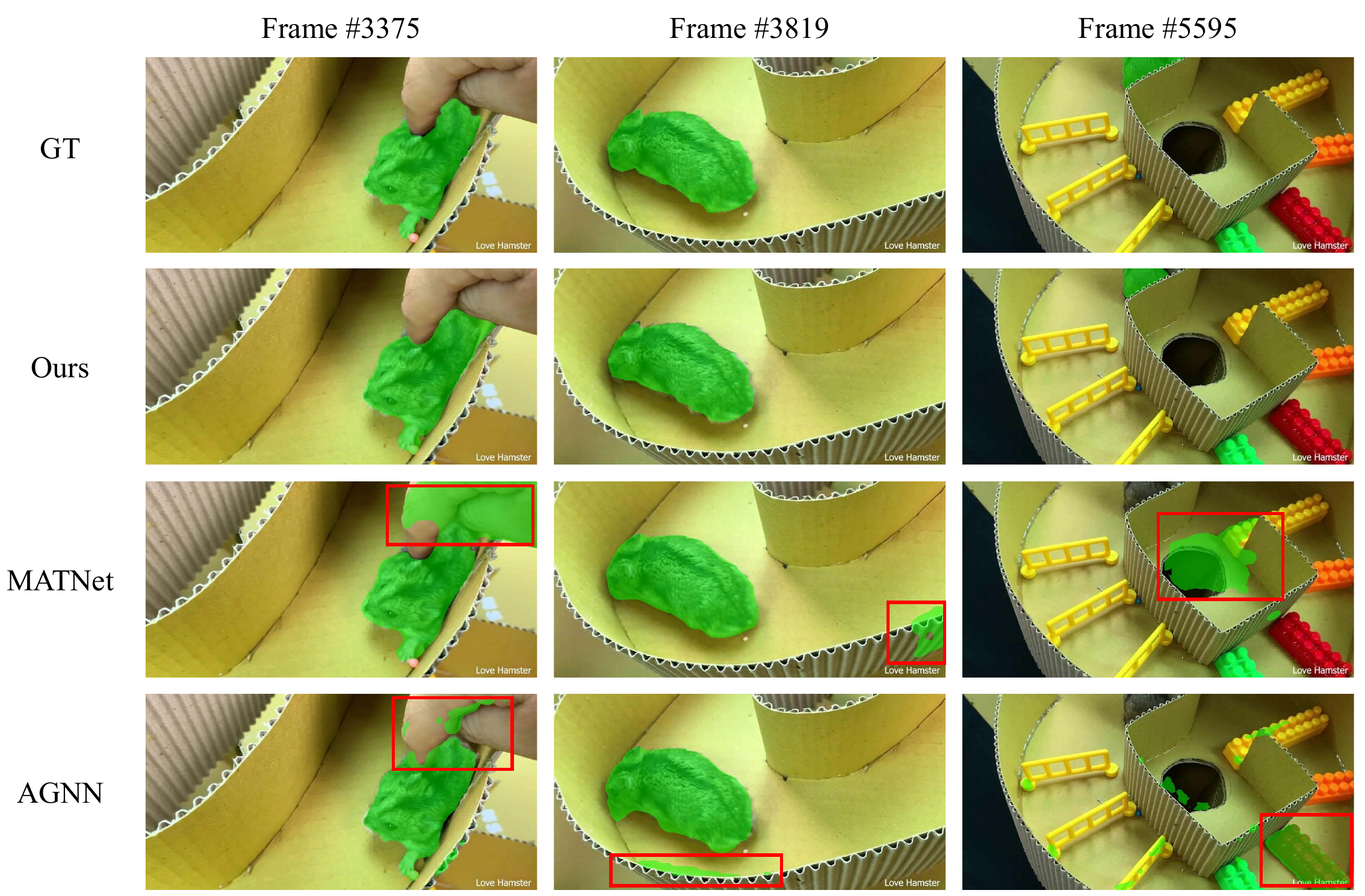}}
	\caption{Qualitative comparisons of videos, (a) \textit{breakdance} (dynamic background) and (b) \textit{scooter-black} (fast motion) from the DAVIS-16~\cite{perazzi2016benchmark} val-set. (c) \textit{train0024} (Slow-moving) from the YouTube-Objects~\cite{prest2012learning} test-set. (d) \textit{rat} (long-term video, scale variations and occlusion) from the Long-Videos~\cite{liang2020video} val-set.}
	\label{fig.6}
\end{figure*}

\begin{table*}[t]
	\centering
	\caption{Quantitative evaluation on the validation set of DAVIS-17~\cite{pont2017} for instance-level ZS-VOS  (see \S{\ref{subsec:5.3}} for details).}\label{Tab.4}
		\begin{tabular}{r|c|c|cc|cc|c}
			\hline
			\multirow{2}*{Method~~~~~~~} &\multirow{2}*{Publication} &\multirow{2}*{Backbone}&\multicolumn{2}{c|}{$\mathcal{J}$} &\multicolumn{2}{c|}{$\mathcal{F}$} &\multirow{2}*{$\mathcal{J}$\&$\mathcal{F}$ Mean$\uparrow$}\\
			~ & ~ & &Mean$\uparrow$ &Recall $\uparrow$ &Mean$\uparrow$ &Recall$\uparrow$ & \\
			\hline
			RVOS~\cite{ventura2019rvos}  &CVPR 2019 &ResNet-101 &36.8 &40.2 &45.7 &46.4 &41.2 \\
			PDB~\cite{song2018pyramid}  &ECCV 2018 &ResNet-50 &53.2 &58.9 &57.0 &60.2 &55.1 \\
			AGS~\cite{wang2019learning}  &TPAMI 2021 &ResNet-101 &55.5 &61.6 &59.5 &62.8 &57.5 \\
			ALBA~\cite{gowda2020alba}  &BMVC 2020 &- &56.6 &63.4 &60.2 &63.1 &58.4 \\
			MATNet~\cite{zhou2020matnet_tip} &TIP 2020 &ResNet-101 &56.7 &65.2 &60.4 &68.2 &58.6 \\
			STEm-Seg~\cite{athar2020stem} &ECCV 2020 &ResNet-101 &61.5 &70.4 &67.8 &\textcolor{green}{75.5} &64.7 \\
			TAODA~\cite{zhou2021target} &CVPR 2021 &ResNet-101 &63.7 &\textcolor{green}{71.9} &66.2 &73.1  &65.0 \\
			UnOVOST~\cite{luiten2020unovost}  &WACV 2020 &ResNet-101 &\textcolor{blue}{66.4} &\textcolor{red}{76.4} &\textcolor{green}{69.3} &\textcolor{blue}{76.9} &\textcolor{green}{67.9} \\
			ProposeReduce~\cite{PRVIS} &ICCV 2021 &ResNet-101 &\textcolor{green}{65.0} &- &\textcolor{red}{71.6} &- &\textcolor{blue}{68.3} \\
                D$^{2}$Conv3D~\cite{schmidt2022d2conv3d} &WACV 2022 &ResNet-101 &60.8 &- &68.5 &- &64.6 \\
			\hline
			\textbf{Ours}  &- &ResNet-101 &\textcolor{red}{67.0} &\textcolor{blue}{75.1} &\textcolor{blue}{71.0} &\textcolor{red}{77.5} &\textcolor{red}{69.0} \\
			\hline
	\end{tabular}
\end{table*}

\section{Experiments}\label{Sec5}
\subsection{Experimental Setup}
\label{Sec5.1}

\subsubsection{Datasets} 
We evaluate HGPU on four publicly available datasets with object-level and instance-level tasks. Specifically, we used DAVIS-16~\cite{perazzi2016benchmark}, DAVIS-17~\cite{pont2017}, YouTube-Objects~\cite{prest2012learning} and Long-Videos~\cite{liang2020video}. DAVIS-16~\cite{perazzi2016benchmark} contains 50 videos, 30 for training and 20 for validation, providing high-quality pixel-level annotations per frame for object-level segmentation. DAVIS-17~\cite{pont2017} is an extended version of DAVIS-16 with 120 video sequences. These sequences are split into 60 training videos, 30 validation videos, and 30 test-dev videos, providing pixel-level object segmentation annotations for instance-based tasks. YouTube-Objects~\cite{prest2012learning} contains 126 web videos from 10 categories, including a total of 20,000 video frames. To evaluate the proposed method on long-sequence videos, we conduct experiments on Long-Videos~\cite{liang2020video}.

\subsubsection{Metrics}
We employs the three standard evaluation metrics recommended by the DAVIS-Benchmark~\cite{perazzi2016benchmark}, namely, region similarity $\mathcal{J}$, boundary accuracy $\mathcal{F}$, and time stability $\mathcal{T}$. For the object-level segmentation datasets, including DAVIS-16, YouTube-Objects, and Long-Videos, the official Matlab code~\cite{perazzi2016benchmark} is used. As for the instance-level dataset, the evaluation framework from~\cite{pont2017} is adopted.

\subsubsection{Reproducibility}
HGPU is implemented using PyTorch and trained on a workstation equipped with an Intel Xeon Gold 5118 CPU (2.30 GHz) and four NVIDIA Tesla V100 32G GPUs. The network is trained with the binary cross-entropy loss function, and online data augmentation is performed by horizontally flipping and rotating the training data within the range of $(-10^{\circ},10^{\circ})$, as presented in this paper. The parameters are optimized using the SGD optimizer, with a weight decay of 1e-5, and the learning rates for the encoder and decoder networks are set to 1e-3 and 1e-2, respectively. The mini-batch size is set to 20, and the entire training process requires approximately two days to complete for 30 epochs.

\subsection{Results for Object-level ZS-VOS}\label{subsec:5.2}

\subsubsection{DAVIS-16 val-set}
Table \ref{Tab.1} presents the quantitative results obtained by comparing our model with state-of-the-art methods on the DAVIS-16 dataset~\cite{perazzi2016benchmark}. HGPU outperforms all existing methods by a significant margin on the DAVIS-16 validation set, with an improvement of 0.5\% in $\mathcal{J}$\&$\mathcal{F}$ Mean compared to the second-best method HCPN~\cite{pei2023HCPN}. Additionally, HGPU results in an increase of 0.2\% and 0.8\% in $\mathcal{J}$ Mean and $\mathcal{F}$ Mean, respectively. When compared to MATNet~\cite{zhou2020matnet_tip}, which employs the same encoder-decoder structure and performs additional pre-training on YouTube-VOS~\cite{xu2018youtube}, HGPU exhibited a significant performance enhancement of 3.6\% and 5.5\% in $\mathcal{J}$ Mean and $\mathcal{F}$ Mean, respectively. Furthermore, when compared to RTNet~\cite{RTNet}, we achieved a 0.9\% improvement on $\mathcal{J}$\&$\mathcal{F}$ while using only half of its FLOPs (93.5G vs. 180.7G).

\subsubsection{YouTube-Objects test-set}
To demonstrate the adaptability of the proposed method to other object-level segmentation datasets, we conduct validation experiments on the YouTube-Objects~\cite{prest2012learning} test-set without further fine-tuning on its training set. Table \ref{Tab.2} gives the detailed results of 10 categories in this dataset. HGPU does not achieve high performance across all categories but has better stability than other methods. Our method outperforms the second-best TMO~\cite{TMO} by 2.4$\%$ in average $\mathcal{J}$ Mean. Quantitative results also prove its versatility for different categories of video segmentation tasks.

\subsubsection{Long-Videos val-set}
The aforementioned DAVIS (about 60 frames per video sequence) and YouTube-Objects (around 150 frames per video sequence) only contain short-term video clips. In contrast, real-world videos tend to have more frames. To verify the performance of our HGPU in long-term VOS, we evaluate it on the Long-Videos~\cite{liang2020video} val-set (2470 frames per video sequence in average). Table~\ref{Tab.3} shows the results under two types of supervision, OS-VOS and ZS-VOS. Our HGPU performs the best, achieving 76.9$\%$ $\mathcal{J}$\&$\mathcal{F}$ Mean under the ZS-VOS supervision. Compared to the second-best method AGNN~\cite{wang2019zero}, our model obtains an improvement of 8.4\%. Meanwhile, HGPU achieves similar results to OS-VOS methods in terms of $\mathcal{F}$ Mean and $\mathcal{F}$ Recall, and obtains the best performance in terms of $\mathcal{J}$ and $\mathcal{F}$ Decay.

\subsubsection{Attitude-based Analysis}
To further verify the performance of the proposed method for various video challenge tasks, we select the DAVIS-16 val-set for the attribute-based analysis. We take a total of 15 most influential attributes~\cite{perazzi2016benchmark}, where each video sequence has one or more attributes and each attribute represents a specific type of video challenge. We present the attribute-based results of three brightly performing ZS-VOS methods (\ie, MATNet \cite{zhou2020matnet_tip}, DFNet \cite{zhen2020learning} and 3DCSeg \cite{mahadevan2020making}) and our HGPU, as shown in Fig. \ref{fig.5}. 

HGPU achieves first place in all 15 attribute-based video tasks in $\mathcal{J}$ Mean, which significantly outperforms all comparison methods in multiple attributes (\eg, CS, DE, EA, IO, MB and SV). Our approach also obtains leading performance in the majority of attributes over $\mathcal{F}$ Mean. The attribute-based analysis demonstrates the robustness of our method and its ability to meet a wide range of challenges.

\subsubsection{Qualitative Comparisons for Object-level ZS-VOS}
Qualitative comparisons of object-level ZS-VOS tasks is given in Fig.~\ref{fig.6}. We select the official precalculated segmentation masks of the current state-of-the-art methods (\ie, AGNN~\cite{wang2019zero}, 3DCSeg~\cite{mahadevan2020making}, MATNet~\cite{zhou2020matnet_tip}, UnOVOST~\cite{luiten2020unovost}) on three datasets: DAVIS-16~\cite{perazzi2016benchmark}, YouTube-Objects~\cite{prest2012learning}, and Long-Videos~\cite{liang2020video}. Mis-segmentation cases in Fig.~\ref{fig.6} are highlighted using red bounding boxes. Our proposed method, HGPU, outperforms the existing methods in all four video sequences. The visualizations demonstrate the generalizability of our approach.

\subsection{Results for Instance-level ZS-VOS}\label{subsec:5.3}
\subsubsection{DAVIS-17 val-set}
Instance-level ZS-VOS requires predicting pixel-wise instance segmentation, making it more challenging than object-level ZS-VOS. 
We perform experiments on the instance-level dataset DAVIS-17~\cite{pont2017}. Table \ref{Tab.4} reports the quantitative performance of the eight state-of-the-art methods and HGPU. 
The results show that our method outperforms all existing methods. 
For instance, HGPU brings an improvement of 2.0\% on $\mathcal{J}$ Mean compared to the second-best method, ProposeReduce~\cite{PRVIS}. Compared to TAODA~\cite{zhou2021target}, HGPU provides improvements of 3.3\% and 4.8\% in terms of $\mathcal{J}$ Mean and $\mathcal{F}$ Mean, respectively. Experimental results on DAVIS-17 verify the scalability of the proposed method.

\subsubsection{Qualitative Comparisons for Instance-level ZS-VOS}
\begin{figure}[t]
	\begin{center}
		\includegraphics[width=\linewidth]{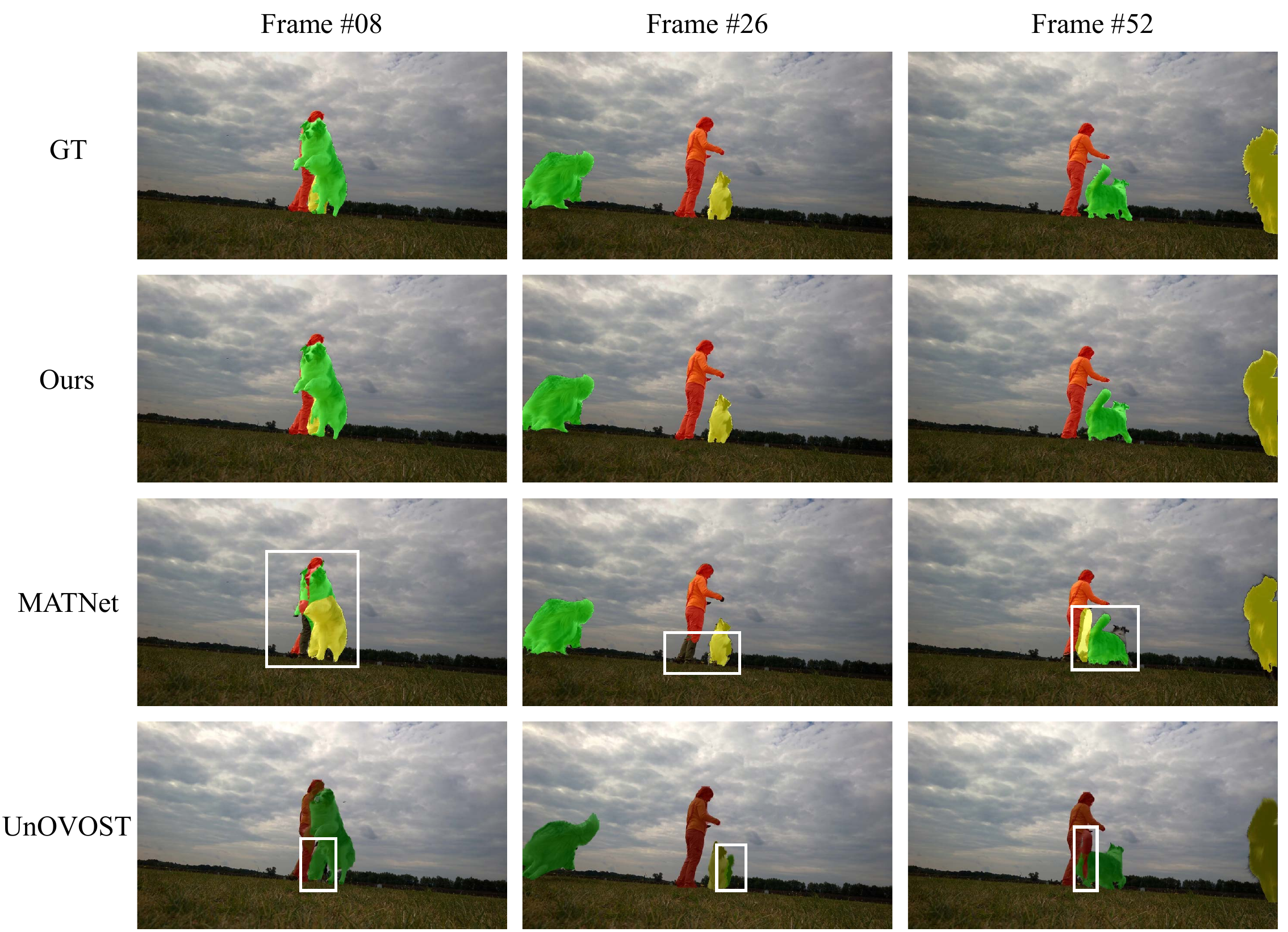}
	\end{center}
	\caption{Qualitative comparisons of the video sequence \textit{dogs-jump} from the DAVIS-17~\cite{pont2017} val-set. In the presence of object interaction between dogs and a woman, instance-level segmentation results are required to be generated.}
	\label{fig.7}
\end{figure}

We provide a qualitative comparison of instance-level ZS-VOS tasks. Fig.~\ref{fig.7} shows the results of our selected top-performing comparison methods on the \textit{dogs-jump} video from DAVIS-17~\cite{pont2017}.  We compare our proposed model to MATNet \cite{zhou2020matnet_tip} and UnOVOST \cite{luiten2020unovost}. Mis-segmentation cases are highlighted using white bounding boxes. Our proposed model gives good segmentation results for this case with challenges such as fast motion, occlusion, and interaction of objects. We attribute this to the fact that the higher-order feature representations of appearance and motion built by the HGPU are suitable for discriminating complex primary object regions.

\begin{figure}[t]
	\begin{center}
		\includegraphics[width=\linewidth]{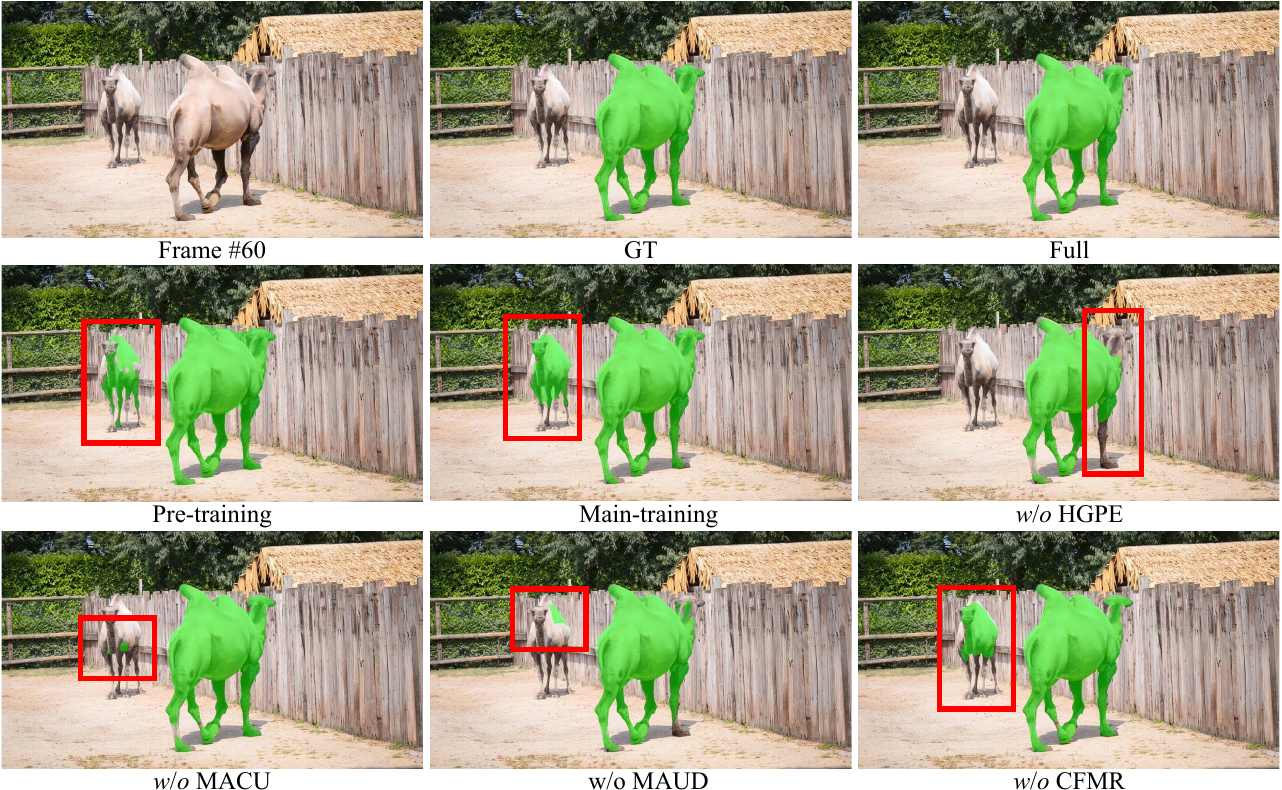}
	\end{center}
	\caption{Ablation study on video sequence \textit{camel} from the DAVIS-16~\cite{perazzi2016benchmark} val-set. The foreground camel needs to be segmented from a similar background.}
	\label{fig.8}
\end{figure}

\subsection{Ablation Study}\label{subsec:5.5}
To quantify the effect of each fundamental component in HGPU, we conduct an exhaustive ablation study on the DAVIS-16 val-set. Table~\ref{Tab.5} presents the comparison results of our full model and its ablated versions. In addition, Fig.~\ref{fig.8} shows the qualitative comparison results of ablation modules.

\subsubsection{Impact of Training Procedure} 
The full training of HGPU is divided into two steps: pre-training on YouTube-VOS~\cite{xu2018youtube} and main-training on DAVIS-16~\cite{perazzi2016benchmark}. To evaluate the impact of the two training procedures, we provide the results of pre-training-only and main-training-only in Table \ref{Tab.5}. Notably, our main-training-only model outperforms many existing methods that use extra training data.

\subsubsection{Efficacy of Encoder-Decoder Architecture}
The encoder-decoder architecture of our proposed method has HGPE and MAUD components. To study its effectiveness, we conduct component independence experiments. The ablated model ``\textit{w}{/}\textit{o} HGPE'' degrades by 7.8\% and 10.7\% in terms of $\mathcal{J}$ Mean and $\mathcal{F}$ Mean, while the model ``\textit{w}{/}\textit{o} MAUD'' drops by 5.6\% and 8.6\%. The absence of the HGPE and MAUD modules hurts the performance. This fact effectively demonstrates the indispensability of performing hierarchical graph pattern understanding in improving video segmentation performance.

\begin{figure*}
	\centering
	\includegraphics[width=1.0\linewidth]{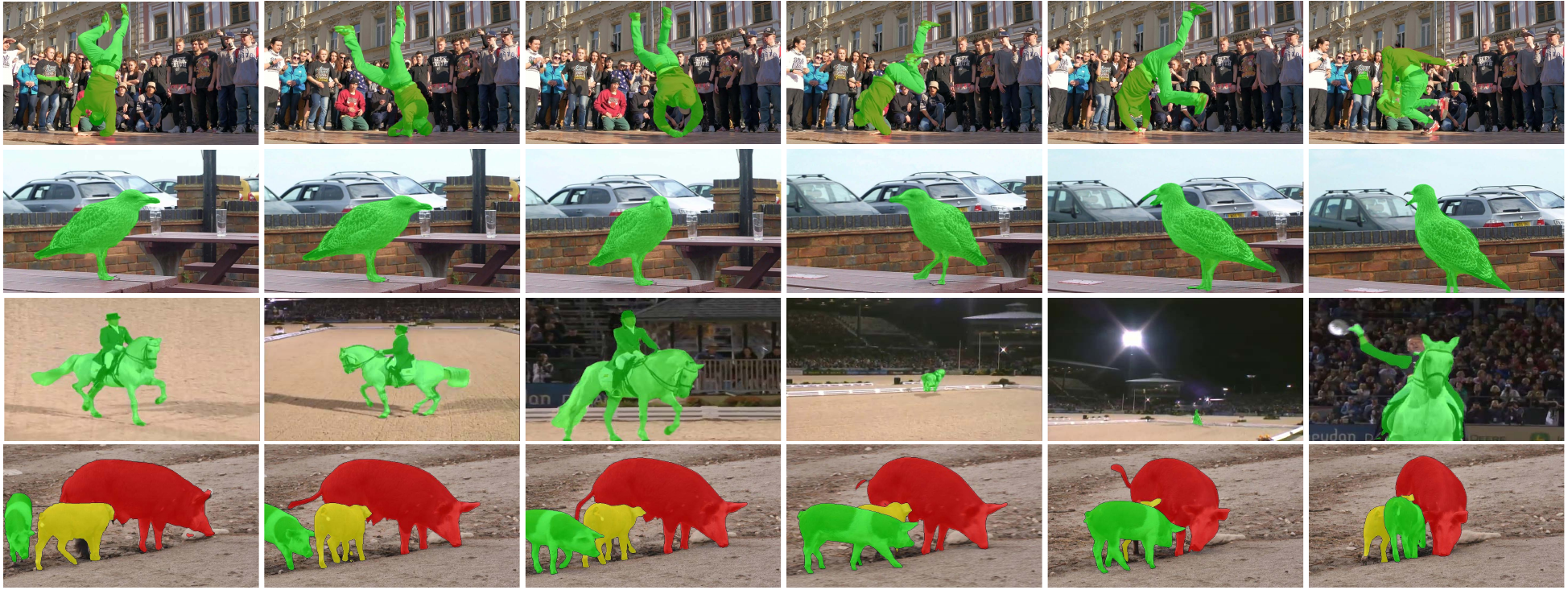}
	\caption{Qualitative results on four videos. From top to bottom: \textit{breakdance} from DAVIS-16 \cite{perazzi2016benchmark}, \textit{bird-0014} from YouTube-Objects \cite{prest2012learning}, \textit{dressage} from Long-Videos \cite{liang2020video}, \textit{pigs} from DAVIS-17 \cite{pont2017}.}\label{fig.9}
\end{figure*}

\subsubsection{Efficacy of Crucial Modules}
When comparing our full HGPU to the ablated models without MAR, CFMR, and MACU, quantitative results reveal an interesting phenomenon. 
\textbf{First}, the structural feature representation of the input video frames can effectively constrain the differences in foreground object features among different frames. 
\textbf{Second}, meaningful pattern recognition and understanding of the motion-appearance features can enhance the interaction between the target object and the edge information throughout the video sequence. 
\textbf{In addition}, the MACU module, which performs motion-appearance feature information transfer between the encoder and decoder, further enriches the motion- and appearance-specific features. 
The performance losses for the ablated models MAR, CFMR, and MACU are 2.0\% (1.8\%), 2.7\% (1.7\%), and 1.2\% (0.9\%) in terms of $\mathcal{J}$ ($\mathcal{F}$) Mean.

\subsubsection{Impact of data augmentation}
Following previous methods~\cite{wang2019learning,AMC-Net,TransportNet,zhou2020matnet_tip}, we have applied online data augmentation techniques such as flipping and rotating on the training datasets. Additionally, we conduct an ablation study to assess the impact of data augmentation on overall performance. Similar to other vision-related tasks, we affirm that data augmentation is also effective for ZS-VOS.

\subsubsection{Impact of optical flow}
We perform an ablation study employing PWC-Net~\cite{sun2018pwc} to investigate the effect of different optical flow estimation methods on the proposed model. 
Although better optical flow estimation usually leads to better segmentation accuracy, our model reveals considerable robustness against the quality of optical flow estimation.

\begin{table}[t]
	\centering
 	\caption{Ablation study on the DAVIS-16~\cite{perazzi2016benchmark} (see \S{\ref{subsec:5.5}} for details).}\label{Tab.5}
		\begin{tabular}{c|c|cc}
			\hline
			\multirow{2}*{Component} &\multirow{2}*{Variant} &\multicolumn{2}{c}{DAVIS-16} \\ 
			~ &~ & Mean &$\Delta$ \\
			\hline
			\multirow{2}*{Reference} &\multirow{2}*{Full HGPU} &\textbf{86.0} ($\mathcal{J}$) &- \\
			& &\textbf{86.2} ($\mathcal{F}$) &- \\
			\hline
			\multirow{4}*{Training} &\multirow{2}*{Pre-training only} &78.6 ($\mathcal{J}$) &-7.4 \\ 
			& &77.8 ($\mathcal{F}$) &-8.4 \\ 
			\cline{2-4}
			~ &\multirow{2}*{Main-training only} &80.3 ($\mathcal{J}$) &-5.7 \\
			& &79.1 ($\mathcal{F}$) &-7.1 \\
			\hline
			\multirow{4}*{Encoder-Decoder}  &\multirow{2}*{\textit{w}{/}\textit{o} HGPE} &78.2 ($\mathcal{J}$) &-7.8 \\
			& &75.5 ($\mathcal{F}$) &-10.7 \\
			\cline{2-4}
			~&\multirow{2}*{\textit{w}{/}\textit{o} MAUD} &80.4 ($\mathcal{J}$) &-5.6 \\
			& &77.6 ($\mathcal{F}$) &-8.6 \\
			\hline
			\multirow{6}*{Module} &\multirow{2}*{\textit{w}{/} HGPE + \textit{w}{/}\textit{o} MAR} &84.0 ($\mathcal{J}$) &-2.0 \\
			& &84.4 ($\mathcal{F}$) &-1.8 \\
			\cline{2-4}
			~&\multirow{2}*{\textit{w}{/} MAUD + \textit{w}{/}\textit{o} CFMR} &83.3 ($\mathcal{J}$) &-2.7 \\
			& &84.5 ($\mathcal{F}$) &-1.7 \\
			\cline{2-4}
			~&\multirow{2}*{\textit{w}{/} MAUD + \textit{w}{/}\textit{o} MACU} &84.8 ($\mathcal{J}$) &-1.2 \\
			& &85.3 ($\mathcal{F}$) &-0.9 \\
			\hline
			\multirow{2}*{Optical Flow} &\multirow{2}*{\textit{w}{/} PWC-Net} &85.2 ($\mathcal{J}$) &-0.8 \\
			& &85.4 ($\mathcal{F}$) &-0.8 \\
			\cline{2-4}
			\hline
                \multirow{2}*{Data Augmentation} &\multirow{2}*{\textit{w}{/}\textit{o} Flipping \& Rotating} &84.9 ($\mathcal{J}$) &-1.1 \\
			& &85.2 ($\mathcal{F}$) &-1.0 \\
                \hline
			\multirow{2}*{Post-Process} &\multirow{2}*{\textit{w}{/}\textit{o} CRF} &85.1 ($\mathcal{J}$) &-0.9 \\
			& &85.3 ($\mathcal{F}$) &-0.9 \\
			\hline
	\end{tabular}
\end{table}

\begin{figure}
	\centering
	\subfloat[car-roundabout]{\includegraphics[width=\linewidth]{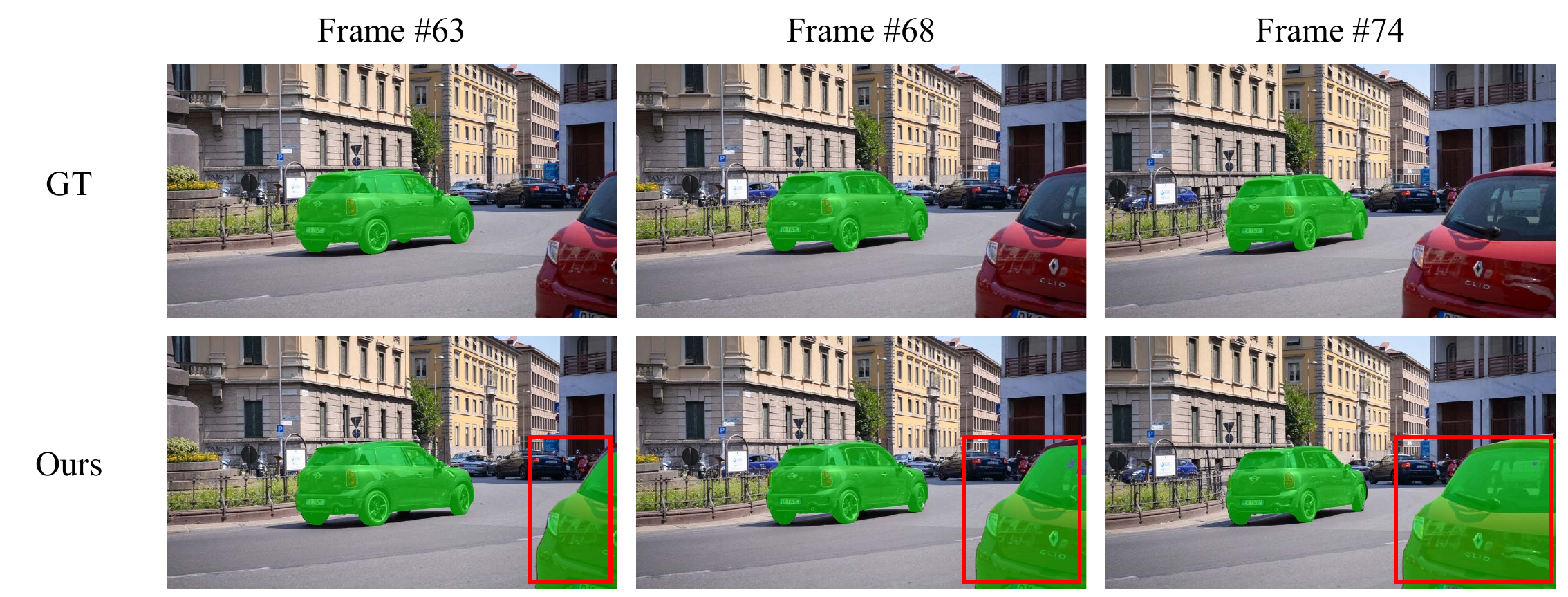}}
	\hfill
	\subfloat[horsejump-high]{\includegraphics[width=\linewidth]{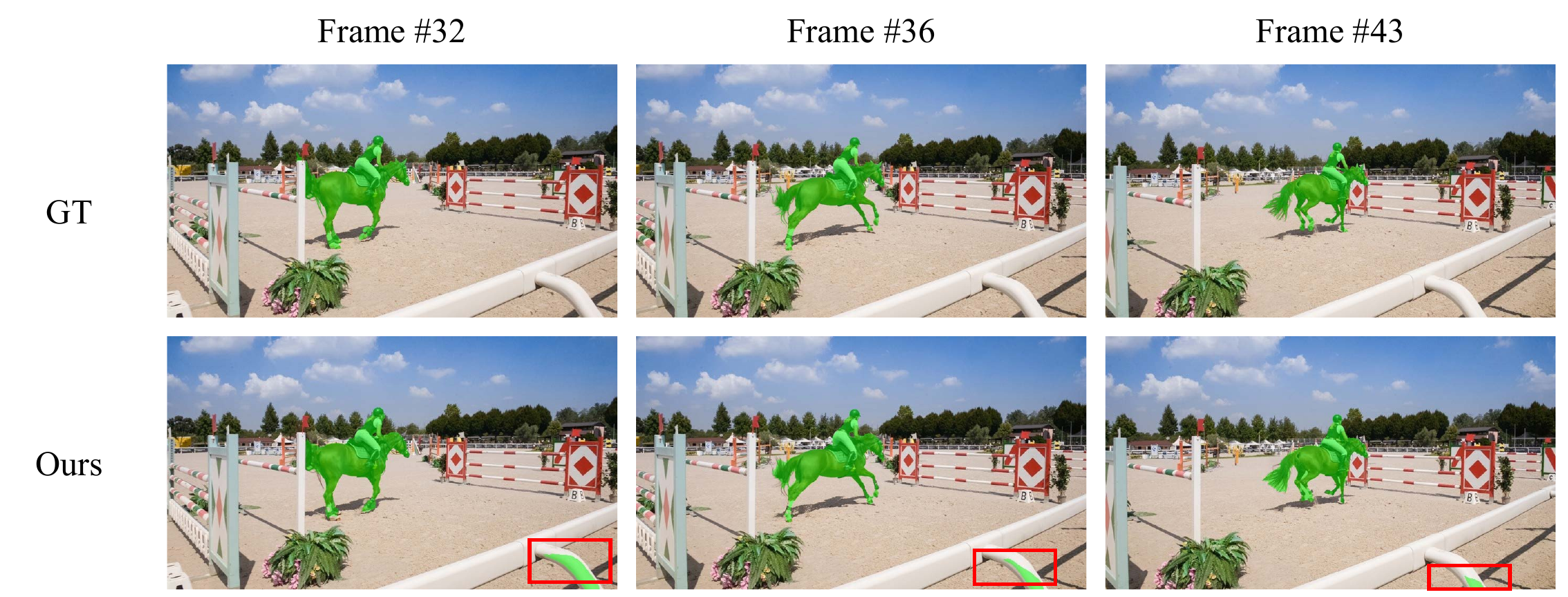}}
	\caption{Qualitative results of the video sequences (a) \textit{car-roundabout} and (b) \textit{horsejump-high} from the DAVIS-16~\cite{perazzi2016benchmark} val-set. We show the segmentation results of our HGPU, where camera shake is present in these videos.}
	\label{fig.10}
\end{figure}

\subsection{Additional Qualitative Results}\label{Sec5.3}
\subsubsection{Success Cases}
Fig. \ref{fig.9} shows the overall qualitative results of HGPU. 
We handpick four videos from the DAVIS-16~\cite{perazzi2016benchmark}, YouTube-Objects~\cite{prest2012learning}, Long-Videos~\cite{liang2020video} and DAVIS-17~\cite{pont2017} validation sets.
These four videos contain multiple challenging frame sequences (\eg, dynamic background, scale variation, interacting objects and occlusion). 
As shown in the top two rows, our method yields desirable results for dynamic, similar, and complex backgrounds. 
In the third and fourth rows, satisfactory segmentation results are acquired under large variation and object interaction.

\subsubsection{Failure Cases}
In Fig.~\ref{fig.10}, we show the two failure cases of our HGPU. 
The limitations of motion-based approaches in VOS tasks are highlighted by failure cases. Mis-segmentation problems can arise, particularly when the target object is in motion in the scene, due to short-term spatio-temporal modeling. The lack of long-term spatio-temporal modeling of the video inter-frame is the root cause of these breakdowns in motion-based methods. Motion-based approaches can easily confuse the target object with other moving objects in the scene, resulting in mis-segmentation, without considering the long-term context. The target object was erroneously identified as the object movement caused by the camera view offset in this case. In summary, long-term spatio-temporal modeling is necessary to improve the accuracy of motion-based approaches in video object segmentation tasks.

\section{Conclusion}
We present HGPU, a hierarchical graph model for learning representations of video frames and interpreting high-order relations among objects in a sequence. The model uses the hierarchical graph pattern encoder (HGPE) and the motion-appearance understanding decoder (MAUD) to acquire spatio-temporal \textit{primary objects'} structural patterns. HGPU performs object- and instance-level segmentation on video sequences with scale variation, occlusion, and noisy backgrounds under zero-shot conditions. Our method performs better than current state-of-the-art methods on four ZS-VOS datasets. Multiple ablation experiments confirm the proposed components' effectiveness. HGPU is currently computationally intensive, which limits its real-time performance. Future work should focus on optimizing the model's architecture to improve its efficiency and reduce computation time, and extensions to other video tasks such as action recognition, tracking, and event detection. To summarize, we hope that the representation of structural appearance and motion features introduced in this paper will receive more attention from the academic community.

{\small
	\bibliographystyle{IEEEtran}
	\bibliography{egbib}
}

\vfill

\end{document}